\documentclass{article}
\usepackage[preprint]{corl_2026}

\usepackage{microtype}
\usepackage{graphicx}
\usepackage{subcaption}
\usepackage{algorithm}
\usepackage{algorithmic}
\usepackage{amsmath,amssymb,bm}
\usepackage{booktabs}
\usepackage{multirow}
\usepackage{tabularx}
\usepackage{wrapfig}
\usepackage{needspace}
\usepackage{float}
\usepackage[most]{tcolorbox}

\hypersetup{
  pdftitle={Learning from Hindsight for VLA Reinforcement Learning},
  pdfauthor={Iris Xu, Sunshine Jiang, John Marangola, Pulkit Agrawal, Zhang-Wei Hong},
  pdfsubject={arXiv preprint}
}

\title{Learning from Hindsight for VLA Reinforcement Learning}
\author{%
\makebox[\textwidth][c]{%
\begin{minipage}{\textwidth}
\centering\small
Iris Xu\textsuperscript{1},
Sunshine Jiang\textsuperscript{1},
John Marangola\textsuperscript{1},\\[0.2em]
Pulkit Agrawal\textsuperscript{1},
Zhang-Wei Hong\textsuperscript{2}\\[0.5em]
{\footnotesize
\textsuperscript{1}Massachusetts Institute of Technology \quad
\textsuperscript{2}MIT-IBM Watson AI Lab
}
\end{minipage}%
}%
}

\begin{document}
\maketitle
\begingroup
\renewcommand{\thefootnote}{}
\footnotetext{\scriptsize Preprint. Correspondence: \texttt{irisxu@mit.edu}}
\endgroup

\begin{abstract}

Reinforcement learning is increasingly used to fine-tune vision-language-action (VLA) models, but robot interaction is expensive and learning becomes highly sample inefficient when successful rollouts are rare. When reward is assigned only for completing the commanded task, a failed rollout is treated as having no value even if it successfully executes behaviors relevant to that task. A robot that fails to place the correct object in a bowl may still move that object toward the bowl or place a different object inside it, demonstrating objects and actions that can be reused to solve the target task. These behaviors define auxiliary tasks that the policy can already solve, providing useful learning signals even before it can solve the harder target task. We introduce \textit{Learning from Hindsight (LfH)}, which turns such failures into additional learning signals. Using a pretrained vision-language model, LfH relabels failed rollouts with the behaviors they actually accomplish and trains the policy jointly on the commanded task and these auxiliary tasks. On out-of-distribution LIBERO-PRO manipulation tasks, LfH matches the final performance of GRPO with approximately \(5\times\) fewer rollouts and improves sample efficiency across multiple VLA backbones. On a physical Franka robot, LfH raises success from \(0\%\) to \(56\%\) within 160 training rollouts, while GRPO reaches \(22\%\).
\end{abstract}

\section{Introduction}
\label{sec:intro}

Reinforcement learning (RL) \cite{kaelbling1996reinforcement} is increasingly used to fine-tune vision-language-action (VLA) models for robot control \cite{black2024pi0,chen2025pirl}. A typical pipeline adapts a VLA to a new task by supervised fine-tuning (SFT) on a few teleoperated demonstrations, then continues with RL so that the robot improves from its own interaction. But when the initial policy rarely succeeds, most rollouts receive no reward and contribute little to learning, making RL fine-tuning highly sample inefficient.

\begin{figure}[!t]
    \centering
    \includegraphics[width=\textwidth]{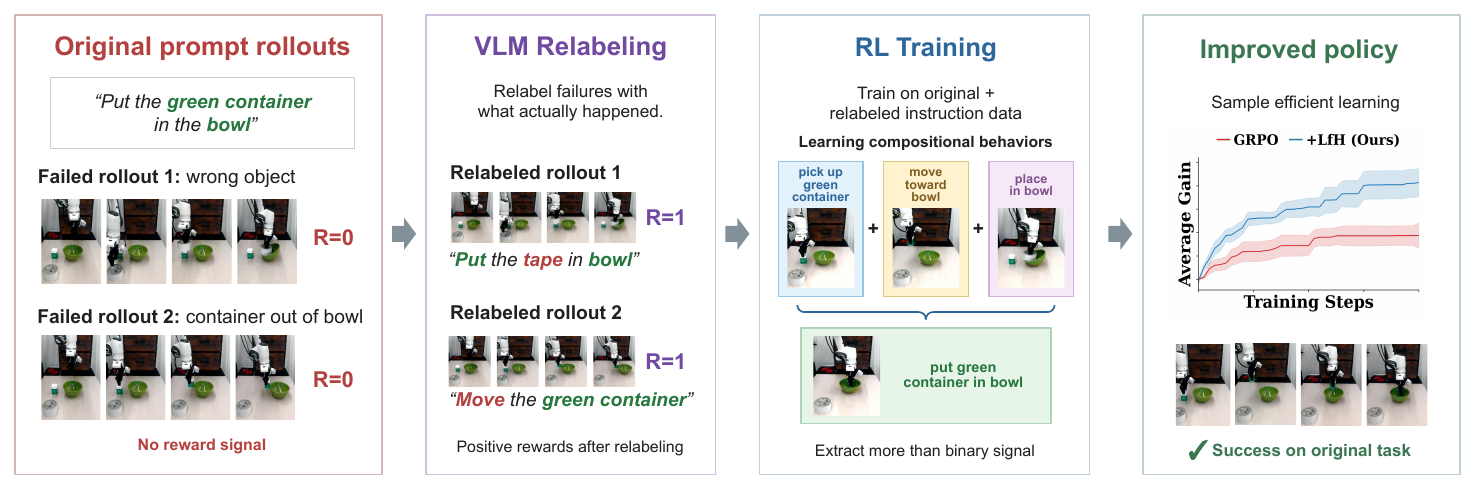}
    \caption{Overview of Learning from Hindsight (LfH). Both rollouts fail the target task ``Put the green container in the bowl'' and receive zero reward, yet each completes a related task, placing the tape in the bowl or moving the green container. A VLM relabels each rollout with the task it actually accomplished, turning both failures into successes. Training on these hindsight tasks with RL teaches skills that the policy composes to solve the target task with fewer samples.}
    \label{fig:teaser}
\end{figure}

A failed rollout, however, is not necessarily useless data. In Figure~\ref{fig:teaser}, every attempt at ``Put the green container in the bowl'' fails, yet one rollout puts the tape in the bowl and another moves the green container toward it. Each rollout succeeds at some task, so relabeling it with the instruction it did satisfy turns a failure into a success. These hindsight tasks are also building blocks of the target task: one placing an object in the bowl and the other moving the green container. Because VLAs generalize compositionally over familiar objects and skills~\cite{brohan2022rt1,brohan2023rt2}, learning these building blocks helps the policy put them together.

This strategy, known as hindsight relabeling, is well established in goal-conditioned RL. Hindsight experience replay \cite{andrychowicz2017hindsight} relabels a trajectory that misses its goal with the outcome it actually reached, and later work extends the idea to richer goal spaces such as images \cite{rauber2017hindsight,pathak2018zero,eysenbach2022contrastive,sahni2019visual}. In these settings, relabeling is straightforward, because the outcome a trajectory reached is directly observable and can serve as its goal. For a language-conditioned VLA, however, the instruction a rollout satisfied is not observed and must be inferred from the robot's behavior. Prior language relabeling for robots operates offline, deriving new instructions for fixed datasets from existing annotations or primitive behaviors \cite{zhang2024sprint,glossop2025cast,xiao2023dial}. Inferring what a failed rollout accomplished while the policy is still being trained remains an open problem.

Pretrained vision-language models (VLMs) \cite{yang2025qwen3} offer a natural way to infer what a robot accomplished. Given a rollout video, a VLM can describe the robot's behavior in open-ended language, and this description can replace the rollout's original instruction. Using these labels for RL fine-tuning, however, raises two challenges. First, a VLM asked to describe a rollout will usually produce some instruction, even when the robot only moved accidentally, and its labels can be poorly grounded in the scene. Second, relabeled rollouts are off-policy, because the policy produced them while following a different instruction, so the update must correct for this mismatch.

We propose Learning from Hindsight (LfH), which relabels failed rollouts with a VLM during RL fine-tuning and addresses both challenges. For the first, LfH treats relabeling as a constrained judgment rather than free-form captioning. The VLM is grounded in the objects present in the scene and may answer that a rollout accomplished nothing. A separate query then checks whether rollouts actually complete the hindsight instruction and may answer unsure when the outcome is ambiguous, so a label that does not match the behavior earns little or no reward. In our ablation on LIBERO-PRO Goal, removing any of these options reduces the gains of LfH (Section~\ref{app:ablation}). For the second, the exact correction \cite{rauber2017hindsight} multiplies probability ratios across every step of the trajectory, a product that vanishes or explodes over long VLA horizons. LfH instead corrects only for the state and action at each step, accepting a small bias in exchange for much lower variance (Section~\ref{app:additional_analysis}). 
 
Our principal contribution is showing that VLMs can turn failed rollouts into useful training data for VLAs. On LIBERO-PRO \cite{zhou2025liberopro} manipulation tasks whose instructions differ from those seen in demonstrations, LfH matches the final performance of Group Relative Policy Optimization (GRPO) \cite{shao2024deepseekmath} with approximately \(5\times\) fewer rollouts. It also outperforms learned progress rewards \cite{ma2024vision,liang2026robometer}, which measure progress only on the commanded task. These gains hold across VLA backbones and extend to a real Franka robot, where, within 160 training rollouts, LfH raises success from \(0\%\) to \(56\%\) while GRPO reaches \(22\%\).

\section{Related Work}

\textbf{Hindsight and Language Relabeling.} Hindsight Experience Replay (HER)~\cite{andrychowicz2017hindsight} relabels a trajectory with the goal it actually achieved, and later work extends the idea to image goals~\cite{sahni2019visual}. Language relabeling has mostly been used to enlarge offline datasets. Zhang et al.~\cite{zhang2025learning} relabel agent trajectories in the Craftax game with an LLM, DIAL~\cite{xiao2023dial} and SPRINT~\cite{zhang2024sprint} relabel robot datasets for policy learning and pretraining, and CAST~\cite{glossop2025cast} and InstructVLA~\cite{yang2025instructvla} augment VLA training with counterfactual instructions and multimodal instruction tuning. SOAR~\cite{zhou2024autonomousimprovementinstructionfollowing} is closest in applying a VLM to autonomously collected robot data, where it proposes tasks and detects success, but it improves the policy with goal-conditioned behavior cloning rather than RL. In these works, relabeling adds supervision to existing data to broaden the range of instructions a policy can follow. LfH uses relabeling for a different purpose. It targets a task the policy cannot yet solve, relabels the policy's own failures on that task during online RL, and trains on the tasks those failures already accomplish so that progress transfers back to the target.

\textbf{Learned Reward Models.} Reward shaping provides learning signal when successes are rare by rewarding intermediate progress~\cite{ngrewardshaping}. Recent work learns such rewards from visual data, as goal- or language-conditioned value functions~\cite{ma2022vip, ma2023liv}, from demonstrations~\cite{ma2024vision, liang2026robometer, zhai2025vision}, or from preference labels~\cite{wang2024rl}. These models refine how a rollout is scored under its commanded instruction. LfH instead changes which instruction the rollout is credited for, which requires inferring what it accomplished and lets it provide signal even when it makes no progress on the commanded task.

\section{Preliminaries}
\label{sec:prelim}
We study RL fine-tuning of VLA policies
\cite{black2024pi0,kim2024openvla}. At the start of each episode, the environment
samples an instruction $g \sim P_g$ that defines a target task and an initial RGB observation $o_0$. A policy
$\pi_\theta$ rolls out actions $a_t \sim \pi_\theta(\cdot \mid o_t,g)$ for $T$ steps,
yielding a trajectory $\tau = (o_0,a_0,\ldots,a_{T-1},o_T)$ that receives a binary
success reward $R(\tau,g)$, equal to $1$ when $\tau$ completes $g$ and $0$ otherwise.
The objective is to maximize expected success
$J(\theta) = \mathbb{E}_{g \sim P_g,\, \tau \sim \pi_\theta(\cdot \mid g)}[R(\tau,g)]$.

We instantiate LfH with group relative policy optimization (GRPO)
\cite{shao2024deepseekmath}. For each instruction $g$, GRPO samples a group of $K$
trajectories $\{\tau_i\}_{i=1}^K$ from $\pi_{\theta_{\mathrm{old}}}$, scores them with
$R_i = R(\tau_i,g)$, and normalizes advantages within the group:
\begin{align}
\label{eq:grpo_adv}
\hat A_i = \frac{R_i-\mu_R}{\sigma_R+\delta},
\qquad
\mu_R = \frac{1}{K}\sum_{j=1}^K R_j,
\end{align}
where $\sigma_R$ is the group standard deviation and $\delta$ is a small constant. With
the importance ratio
\begin{align}
\label{eq:grpo_defs}
r_{i,t}(\theta) &=
\frac{\pi_\theta(a_{i,t}\mid o_{i,t},g)}
{\pi_{\theta_{\mathrm{old}}}(a_{i,t}\mid o_{i,t},g)},
\end{align}
GRPO maximizes
\begin{align}
\mathcal{L}_{\mathrm{GRPO}}(\theta)
&= \mathbb{E}_{g,\{\tau_i\}}\!\left[\frac{1}{K}\sum_{i=1}^K\sum_{t=0}^{T-1}\ell_{i,t}(\theta)\right], \\
\ell_{i,t}(\theta)
&= \min\!\left(r_{i,t}\hat A_i,\,\bar r_{i,t}\hat A_i\right) - \beta\, \mathrm{KL}\!\left(\pi_\theta \,\|\, \pi_{\mathrm{ref}}\right), \label{eq:grpo}
\end{align}
where $\bar r_{i,t} = \mathrm{clip}(r_{i,t},1-\epsilon,1+\epsilon)$ and $\beta$ controls the KL penalty relative to the SFT checkpoint $\pi_{\mathrm{ref}}$.

\section{Method: Learning from Hindsight (LfH)}
\label{sec:method}

\textbf{Problem.}
A GRPO group provides no learning signal when all its trajectories receive the
same reward. A weak initial policy fails most rollouts, so many groups receive all-zero rewards. These groups have zero normalized advantage in Eq.~\eqref{eq:grpo_adv}, contribute no policy gradient, and fine-tuning makes little progress. LfH recovers learning signal from these otherwise
discarded trajectories.

\textbf{Approach.}
A trajectory that fails under one instruction may succeed under another. In Figure~\ref{fig:teaser}, a robot commanded to ``put the green container in the bowl'' instead places the tape in the bowl. This rollout fails for $g=\text{``put the green container in the bowl''}$ but succeeds for the hindsight instruction $g'=\text{``put the tape in the bowl''}$. LfH relabels failed trajectories with instructions
describing what the robot actually accomplished, converting zero-reward samples
$(\tau,g)$ into successful examples $(\tau,g')$. The hindsight instruction $g'$ need not be drawn from $P_g$. It shares objects and skills with the commanded task, which the policy can later compose to solve that task, and in the low-success regime it provides signal where the commanded instruction gives none. LfH lets the
policy learn from what it already manages to do before it can reliably solve the
commanded tasks. Two questions remain: (1) how to generate hindsight-labeled data
(Section~\ref{subsec:method:her}), and (2) how to incorporate it into GRPO
(Section~\ref{subsec:method:grpo_her}).

\subsection{Training Signals from Hindsight}
\label{subsec:method:her}
For each commanded instruction $g$, GRPO samples a group $\{\tau_i\}_{i=1}^K$ and scores
it with $R_i = R(\tau_i,g)$. LfH activates only on low-reward groups, i.e., those with mean reward $\frac{1}{K}\sum_i R_i < \eta$. High-reward groups already reinforce
the commanded behavior, so relabeling them may introduce redundant or noisy auxiliary supervision. Given a low-reward group, LfH converts it into a
hindsight-labeled group in two steps.

\textbf{Instruction relabeling.}
We infer an alternative instruction under which a failed rollout still provides
supervision. Since robot rollouts may lack privileged object states or task
annotations, the VLM relabeler $\mathcal{M}_\psi$ sees only the rollout video, i.e., RGB frames
from a fixed camera, together with the list of objects $\mathcal{O}$ in the scene. LfH samples an anchor from the failed trajectories,
$i^\star \sim \mathrm{Unif}(\{i:R_i=0\})$, since successful trajectories already
carry a reliable label. Not every failure is useful, because some trajectories contain only accidental motion. To avoid noisy instructions, the VLM first classifies whether the anchor contains meaningful behavior, and anchors labeled uninteresting are discarded. For the remaining anchors, the VLM generates a hindsight instruction $g'$ describing what the anchor did in terms of the objects present in the scene:
\begin{equation}
\label{eq:lfh_instr}
g' \sim \mathcal{M}^{\mathrm{inst}}_\psi(\cdot \mid \tau_{i^\star}, \mathcal{O}).
\end{equation}

\textbf{Reward relabeling.}
LfH then evaluates every trajectory in the group under the shared hindsight instruction $g'$:
\begin{equation}
\label{eq:lfh_reward}
\tilde R_i = \mathcal{M}^{\mathrm{rew}}_\psi(\tau_i,g') \in \{0, 0.5, 1\},
\end{equation}
where $1$ denotes completion of $g'$, $0$ failure, and $0.5$ an ambiguous outcome, so an uncertain judgment is not treated as a confident success or failure.
A shared instruction is essential because GRPO computes advantages relative to other
trajectories in the group, and relabeling each with a different instruction would make their
rewards incomparable. This yields the hindsight group
$\tilde{\mathcal{G}}=\{(\tau_i,g',\tilde R_i)\}_{i=1}^K$.

\subsection{GRPO with Hindsight Relabeling}
\label{subsec:method:grpo_her}
LfH treats the hindsight group as an ordinary GRPO group. The only structural change is that its trajectories sampled under the commanded instruction $g$ are optimized under a hindsight instruction $g'$. The exact hindsight policy gradient (HPG)~\cite{rauber2017hindsight} corrects this shift by multiplying the action-probability ratio $\pi(a_s\mid o_s,g')/\pi(a_s\mid o_s,g)$ over the current and all earlier steps $s\le t$. For long VLA horizons, however, these products vanish or explode as small per-step differences compound. On our rollouts, the median cumulative ratio is $1.9\times10^{-12}$, and $16.6\%$ of cumulative ratios underflow to zero. We therefore use the local per-step ratio:
\begin{equation}
\label{eq:lfh_ratio}
\tilde r_{i,t}(\theta) =
\frac{\pi_\theta(a_{i,t}\mid o_{i,t},g')}
{\pi_{\theta_{\mathrm{old}}}(a_{i,t}\mid o_{i,t},g)}.
\end{equation}
Unlike the GRPO ratio in Eq.~\eqref{eq:grpo_defs}, Eq.~\eqref{eq:lfh_ratio} need not equal one at the start of an update, because the numerator and denominator condition on different instructions. It ignores how the instruction change affects earlier steps, so it is biased, but its variance stays low. The bias is small when $g$ and $g'$ induce similar action distributions, and in practice the median local ratio is $0.94$. We define the hindsight objective $\mathcal{L}_{\mathrm{H\text{-}GRPO}}$ as Eqs.~\eqref{eq:grpo_adv}--\eqref{eq:grpo} with advantages $\tilde A_i$ computed from $\tilde R_i$, the ratio $\tilde r_{i,t}$ in place of $r_{i,t}$, and the KL term evaluated under $g'$. To limit the effect of noisy or mismatched relabels, we clip $\mathcal{L}_{\mathrm{H\text{-}GRPO}}$ before weighting it:
\begin{equation}
\label{eq:lfh_obj}
\mathcal{L}_{\mathrm{LfH}}(\theta) =
\mathcal{L}_{\mathrm{GRPO}}(\theta) +
\lambda\,\min\!\left(\mathcal{L}_{\mathrm{H\text{-}GRPO}}(\theta),c\right).
\end{equation}
Here, $\lambda$ weights the hindsight objective and $c$ is its clipping threshold. We set $c$ near the 85th percentile of early-training hindsight objectives, so the cap binds only on the largest hindsight terms, and choose $\lambda$ so the weighted hindsight term roughly matches the GRPO objective. The original GRPO objective is unchanged. Section~\ref{app:additional_analysis} reports the bias--variance analysis and ratio statistics; Section~\ref{app:mechanism_analysis} gives HPG ablations.

\begin{algorithm}[t]
\caption{Learning from Hindsight (LfH)}
\label{alg:lfh}
\begin{algorithmic}[1]
\REQUIRE Policy $\pi_\theta$, VLM relabeler $\mathcal{M}_\psi$, instruction distribution $P_g$, group size $K$, threshold $\eta$, hindsight weight $\lambda$, clipping threshold $c$
\REPEAT
    \STATE $\mathcal{B} \leftarrow \emptyset$, $\tilde{\mathcal{B}} \leftarrow \emptyset$, $\pi_{\theta_{\mathrm{old}}} \leftarrow \pi_\theta$
    \FOR{each sampled instruction $g \sim P_g$}
        \STATE Roll out $\mathcal{G}=\{(\tau_i,g,R_i)\}_{i=1}^K$ with $\pi_{\theta_{\mathrm{old}}}$; $\mathcal{B} \leftarrow \mathcal{B} \cup \mathcal{G}$
        \IF{$\frac{1}{K}\sum_{i=1}^K R_i < \eta$}
            \STATE Generate hindsight group $\tilde{\mathcal{G}}$ from $\mathcal{G}$ via $\mathcal{M}_\psi$; $\tilde{\mathcal{B}} \leftarrow \tilde{\mathcal{B}} \cup \tilde{\mathcal{G}}$
        \ENDIF
    \ENDFOR
    \STATE Update $\theta$ by maximizing Eq.~\eqref{eq:lfh_obj} on $\mathcal{B}$ and $\tilde{\mathcal{B}}$
\UNTIL{convergence}
\end{algorithmic}
\end{algorithm}

\section{Experiments}
\label{sec:exp}

We organize our experiments around the following questions:
(1) Does LfH improve the sample efficiency of GRPO fine-tuning under sparse rewards? (Sections~\ref{subsec:exp:illustrative} and~\ref{subsec:exp:sample_efficiency}).
(2) How does LfH improve sample efficiency? (Section~\ref{subsec:analysis}).
(3) Is LfH effective on a physical robot? (Section~\ref{subsec:exp:real}). Further results and analyses appear in Section~\ref{sec:additional_results}.

Unless otherwise specified, all methods are initialized from the same \texttt{RLinf-Pi05-LIBERO-SFT} checkpoint~\cite{chen2025pirl}, which was trained with few-shot demonstrations on the four LIBERO task suites~\cite{liu2023libero} in simulation. We implement LfH and the baselines in RLinf~\cite{chen2025pirl,yu2025rlinf}. LfH uses \texttt{Qwen3-VL-235B-A22B-Thinking-FP8} for both instruction relabeling and reward relabeling (Section~\ref{subsec:method:her}); prompts and further examples appear in Sections~\ref{app:lfh_vlm_prompts} and~\ref{app:examples}. Trajectories are represented as sequences of video frames from the global camera facing the robot.

\subsection{Illustrative Example}
\label{subsec:exp:illustrative}

We use the LIBERO-90 task ``close the microwave'' to illustrate how LfH improves sample efficiency and helps RL overcome a cold start from a weak initial policy.

\begin{wrapfigure}{r}{0.60\linewidth}
    \vspace{-6pt}
    \centering
    
    \includegraphics[width=\linewidth]{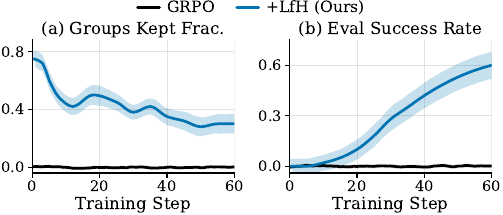}
    
    \vspace{-4pt}
    \caption{\textbf{(a)} Fraction of groups kept during training and \textbf{(b)} evaluation success rate for GRPO and LfH. LfH keeps more trajectory groups through relabeling, while GRPO discards most groups. LfH also escapes the weak initial policy with zero success rate and reaches 60\% success by training step 60.}
    \label{fig:her_illustrative}
    \vspace{-6pt}
\end{wrapfigure}

\textbf{LfH recovers useful training data from failed rollouts.}
The initial policy has zero success on the target task. As a result, standard GRPO assigns nearly all trajectory groups all-zero rewards and discards them, as shown in Figure~\ref{fig:her_illustrative}(a). In contrast, LfH retains nearly 80\% of groups after hindsight relabeling. The remaining groups are discarded when the failures cannot be assigned a meaningful hindsight instruction and reward.

\textbf{LfH improves learning under sparse rewards.}
These recovered training examples allow GRPO to update even before the policy succeeds on the original task. As shown in Figure~\ref{fig:her_illustrative}(b), GRPO+LfH steadily improves on ``close the microwave,'' while standard GRPO remains at zero success.

\textbf{Why can instruction relabeling help?}
Although the policy initially fails at ``close the microwave,'' its rollouts still contain behaviors that can be relabeled with meaningful instructions, such as \emph{``Put the mug inside the microwave''} and \emph{``Pull the microwave door open.''} These examples train the policy on objects and actions shared with the target task, including the \emph{microwave}, the \emph{door}, and the concept \emph{open}, which is the opposite of \emph{close}. We hypothesize that this allows the policy to reuse concepts learned from other relabeled instruction-trajectory pairs when learning the target instruction.

\subsection{Can LfH improve sample efficiency?}
\label{subsec:exp:sample_efficiency}

\textbf{Setup.}
We evaluate whether LfH can accelerate RL fine-tuning when the initial VLA policy is weak. To this end, we construct out-of-distribution (OOD) tasks from the LIBERO-PRO benchmark using its task perturbation. These tasks preserve the original scenes, object instances, and environment dynamics, but change the task specification, such as the instructed object, target state, or desired action. For example, a policy may have seen the same kitchen scene during SFT, but must now manipulate a new object or reach a different goal state. This setting isolates the central challenge of sample-efficient RL: improving from mostly failed rollouts. 

\textbf{Implementation \& Baselines.}
 We train each method separately on each of the four LIBERO-PRO suites (Spatial, Goal, Object, and Long), each containing 10 tasks, with task perturbations. We report results averaged across suites. Each run consists of 40 training steps, collecting 64 environment rollouts per step (8 groups of 8) for Spatial, Goal, and Object, and 512 rollouts per step (64 groups of 8) for Long. We report means and standard errors over four random seeds. We compare LfH with standard GRPO and GRPO with RoboMETER dense progress rewards~\cite{liang2026robometer}. Standard GRPO uses the sparse environment success reward for the original instruction, while GRPO+RoboMETER uses a dense reward measuring progress toward the same instruction. The latter tests whether denser reward feedback alone can account for the sample-efficiency gains of LfH.

\textbf{Metric.}
During training, we periodically freeze the policy and evaluate it with 480 rollouts per suite. We compute normalized gain separately for each task and then macro-average across tasks, giving each task equal weight. Let $\mathrm{SR}_0$ denote the success rate of the initial policy, and let $\mathrm{SR}_t$ denote the evaluation success rate after $t$ training steps. We define
$
\mathrm{Gain}(t)=\frac{\mathrm{SR}_t}{\mathrm{SR}_0}-1.
$
This metric measures how quickly each method improves relative to the initial checkpoint as training proceeds. Every LIBERO-PRO task in this experiment has $\mathrm{SR}_0>0$, so the gain is always defined. The real-world task starts at zero success, so there we report raw success rates.

\begin{figure}[htbp]
    \centering
    \includegraphics[width=0.85\linewidth,trim=0 10bp 0 10bp,clip]{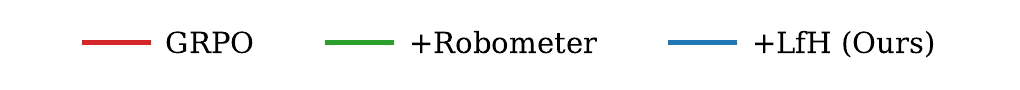}\par\vspace{-2pt}
    \begin{subfigure}[t]{0.32\linewidth}
        \centering
        \makebox[\linewidth][c]{\includegraphics[width=1.08\linewidth]{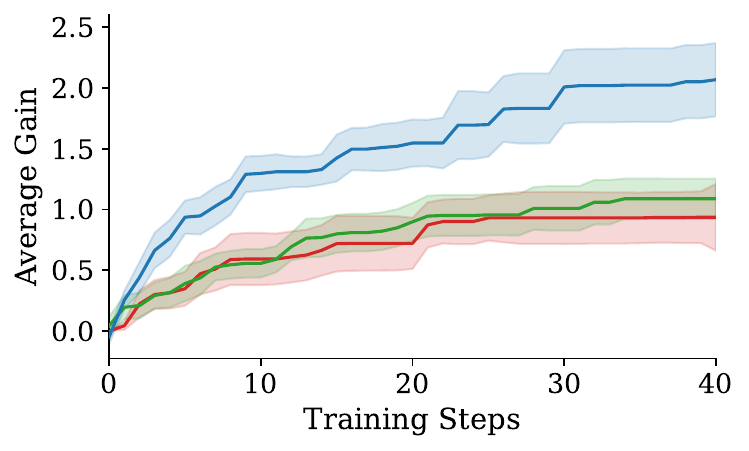}}
        \caption{}
        \label{fig:sample_efficiency}
    \end{subfigure}\hfill
    \begin{subfigure}[t]{0.32\linewidth}
        \centering
        \makebox[\linewidth][c]{\includegraphics[width=1.08\linewidth]{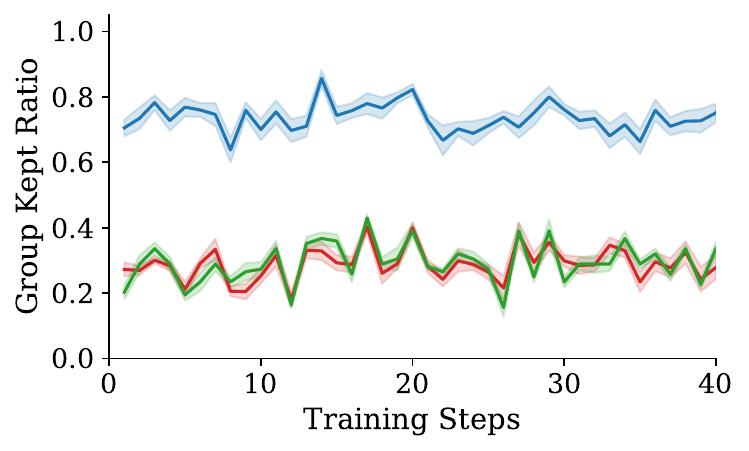}}
        \caption{}
        \label{fig:sample_efficiency_group_kept}
    \end{subfigure}\hfill
    \begin{subfigure}[t]{0.32\linewidth}
        \centering
        \makebox[\linewidth][c]{\includegraphics[width=1.08\linewidth]{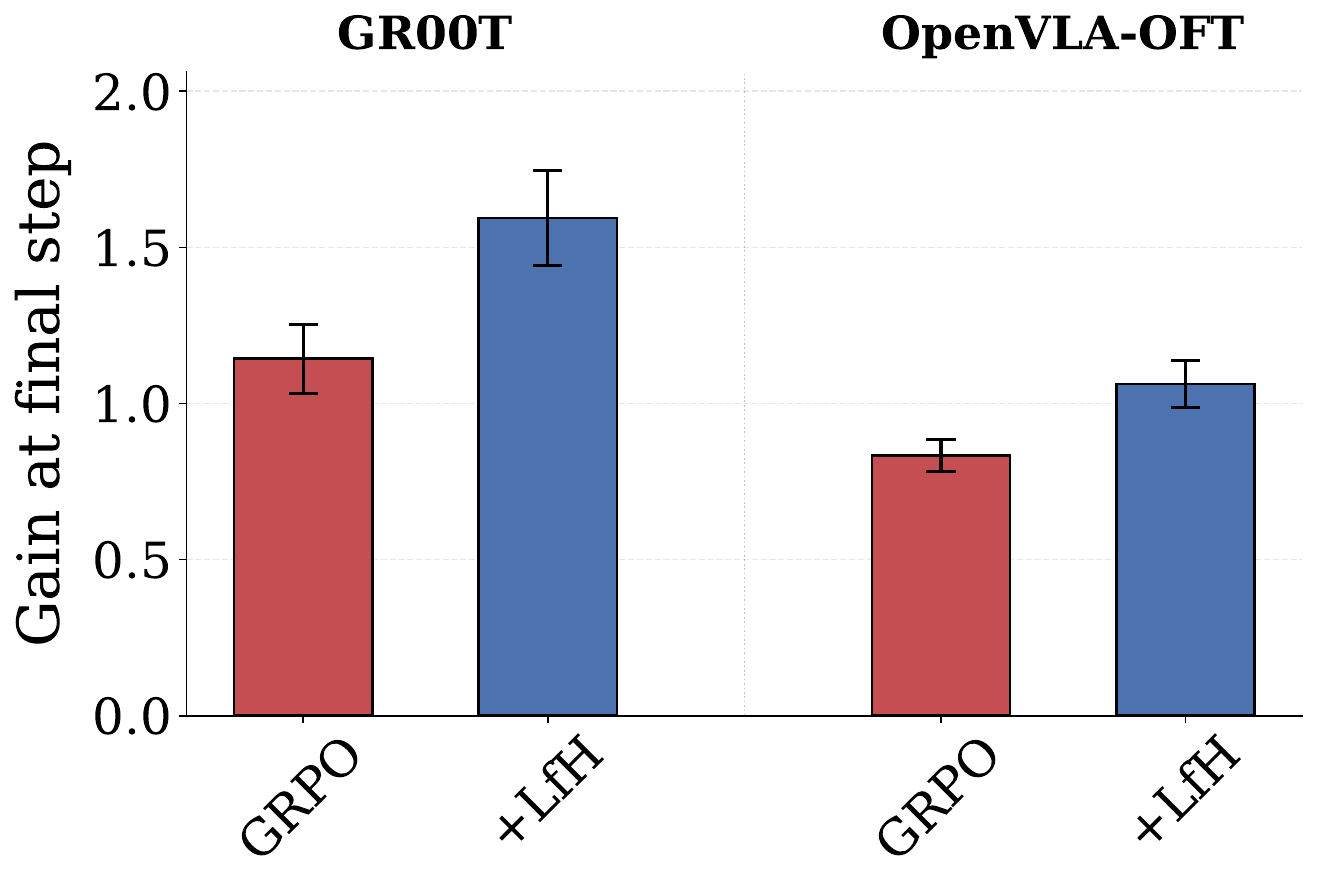}}
        \caption{}
        \label{fig:generality}
    \end{subfigure}
    \caption{ \textbf{(a)} Average success-rate gain during training across LIBERO-PRO suites (task perturbation). \textbf{(b)} Fraction of rollout groups kept during training. LfH uses more groups for training, making better use of rollout data. \textbf{(c)} Final gain on additional VLA backbones shows LfH transfers beyond $\pi_{0.5}$.}
    \label{fig:sample_efficiency_generality}
\end{figure}

\textbf{Results.}
Figure~\ref{fig:sample_efficiency} demonstrates that LfH substantially improves sample efficiency. LfH matches GRPO's plateau performance after only 6 training steps, whereas GRPO requires approximately 30 steps, a $5\times$ reduction in environment interactions. LfH also outperforms GRPO with RoboMETER dense rewards, suggesting that denser feedback alone is insufficient when the policy rarely makes progress toward the commanded instruction. By step 40, LfH outperforms GRPO across all four suites. 

Figure~\ref{fig:sample_efficiency_group_kept} helps explain this gain. Following the default in RLinf~\cite{yu2025rlinf}, groups whose mean reward falls below a lower threshold or above an upper threshold are filtered out before the update, since their near-uniform rewards yield vanishing advantages in Eq.~\eqref{eq:grpo_adv}. Under this filter, GRPO and GRPO+RoboMETER retain only 20--40\% of rollout groups. LfH retains 70--80\%, because relabeling turns many filtered all-failure groups into hindsight groups with informative rewards.

\textbf{Can more updates per rollout match LfH?} We increase the number of update epochs for both GRPO and proximal policy optimization (PPO) \cite{schulman2017ppo}, meaning that each method takes more gradient steps on every rollout batch. The GRPO sweep directly tests whether extra optimization can explain LfH's gain, while PPO provides the same test with another standard RL algorithm. Figure~\ref{fig:ue_sweep} shows that more updates do not consistently help either method and can hurt; even the best variants remain below LfH. Thus, LfH benefits from extracting new supervision from failed trajectories, not simply from reusing the same data more often.

\textbf{Generality.} Finally, we test whether this benefit transfers beyond a single VLA backbone. We keep the suites, task perturbation, rollouts per step, and seeds from the $\pi_{0.5}$ experiments. GR00T~\cite{bjorck2025gr00t} and OpenVLA-OFT~\cite{kim2024openvla} improve more slowly under GRPO, so we train each until its GRPO baseline plateaus, which takes 200 steps for GR00T and 60 steps for OpenVLA-OFT. Figure~\ref{fig:generality} shows consistent gains across both VLAs; per-suite curves appear in Section~\ref{app:full_training_curves}.

\Needspace{15\baselineskip}
\subsection{Analysis: How does LfH improve sample efficiency?}
\label{subsec:analysis}

\begin{wrapfigure}{r}{0.54\linewidth}
    \vspace{-6pt}
    \centering
    \includegraphics[width=\linewidth]{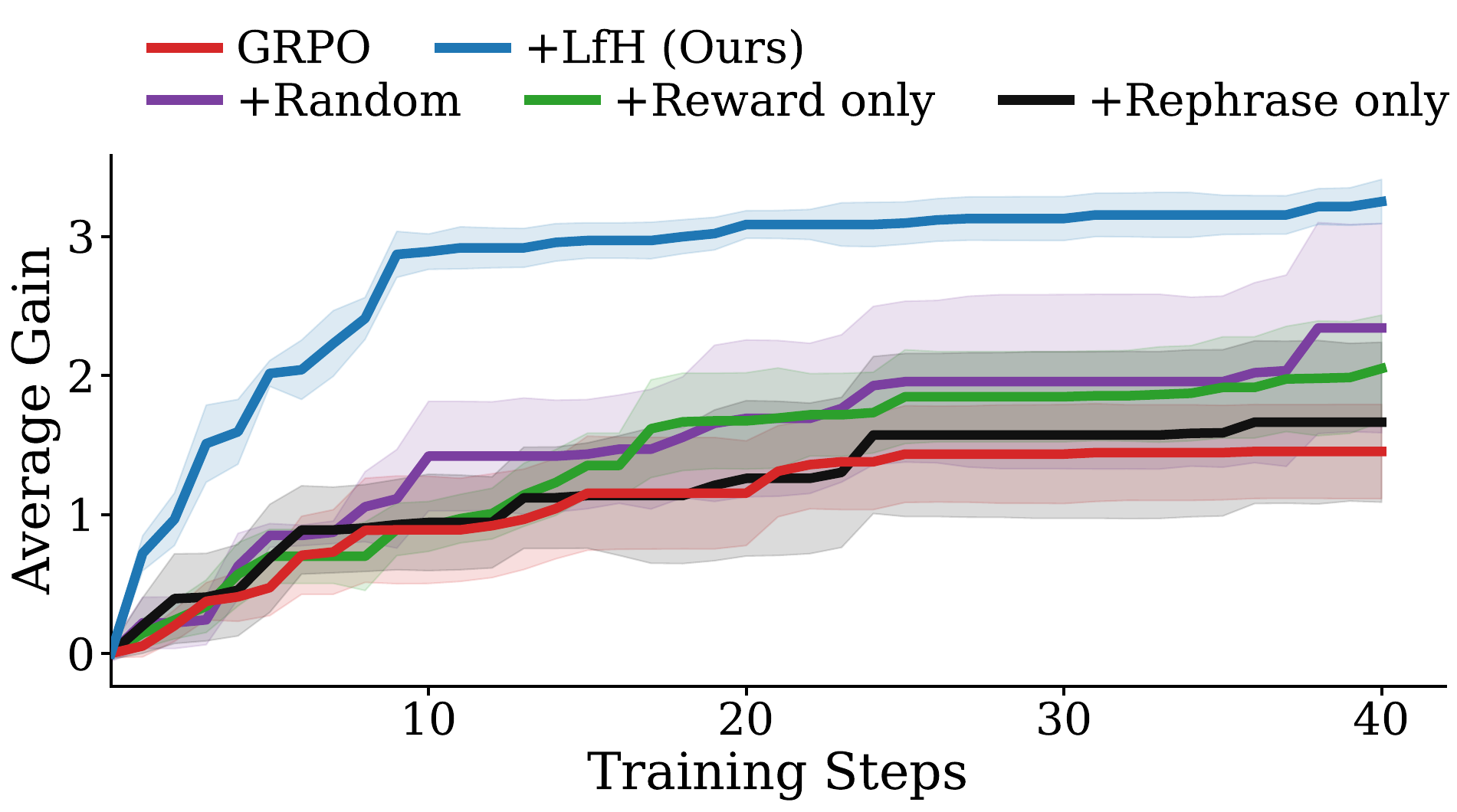}
    \caption{Comparison of LfH with ablations (4 seeds each) that modify parts
    of the hindsight mechanism. LfH outperforms rephrasing,
    reward-only relabeling, and random-reward.}
    \label{fig:ablation_main}
    \vspace{-6pt}
\end{wrapfigure}

\textbf{Can LfH performance gains be attributed to instruction augmentation or reward relabeling alone?}
We isolate the two steps in LfH: hindsight instruction relabeling and reward relabeling (Section~\ref{sec:method}).  We compare against two ablations. \emph{Rephrase-only} replaces the hindsight instruction $g'$ with a paraphrase of the original instruction $g$ while keeping the original rewards, testing whether language augmentation alone explains the gain. \emph{Reward-only} keeps the instruction as $g$ but relabels rewards with the VLM, testing whether reward relabeling alone explains the gain.
Both ablations improve only marginally over GRPO and stay well below LfH throughout training (Figure~\ref{fig:ablation_main}), so the benefit comes from coupling the two steps.

\textbf{Can LfH performance gains be attributed to perturbed rewards?}
We next test whether LfH works simply because it injects reward variance into otherwise discarded groups. The random-reward baseline creates nonzero advantages without semantically meaningful hindsight labels, yet gains little (Figure~\ref{fig:ablation_main}). The benefit of LfH therefore comes from grounded hindsight labels, not from avoiding zero-variance groups.

\textbf{When does LfH help the most?}
LfH is most effective when failures are diverse and semantically interpretable. Figure~\ref{fig:diversity_rollouts} shows sample rollouts at training step 40 on the microwave task. Standard GRPO produces many similar trajectories: the robot repeatedly approaches the same mug with similar gripper poses, leading to a low-diversity failure mode. In contrast, LfH rollouts show more varied object interactions and spatial configurations around the objects. Because these samples come after training has progressed, LfH may also help maintain behavioral diversity during RL fine-tuning, not only exploit diversity the initial policy already had. One possible explanation is that hindsight relabeling gives credit to multiple meaningful alternative behaviors, whereas GRPO treats them all as failures under the commanded task. As a result, GRPO can collapse toward repeated unsuccessful behaviors, while LfH continues to reinforce a broader set of grounded object interactions.

\Needspace{8\baselineskip}
\textbf{How does data diversity affect LfH?}
We vary $\pi_{0.5}$'s rollout noise to control trajectory diversity. Figure~\ref{fig:noise_ablation} shows an optimum for both methods: too little noise limits exploration, while too much produces incoherent rollouts. LfH benefits more near this optimum because it can relabel diverse failures as meaningful alternative tasks, whereas GRPO benefits only when exploration improves reward on the commanded task.

\subsection{Real-world results}
\label{subsec:exp:real}

\textbf{Setup.}
We evaluate LfH on a real Franka FR3 robot with manipulable objects, two drawers, and a two-level rack. We first perform SFT on $\pi_{0.5}$ with 9 language-conditioned manipulation tasks using 20 SpaceMouse demonstrations per task, spanning short pick-and-place and longer multi-stage tasks. We then fine-tune on a held-out task, \emph{put the green container in the bowl}, where the initial SFT policy has zero success. Both GRPO and GRPO+LfH use the same binary human reward. See Section~\ref{app:real_world_details} for details and Section~\ref{app:impl} for implementation settings.

\begin{wrapfigure}{r}{0.44\linewidth}
    \vspace{-6pt}
    \centering
    \includegraphics[width=\linewidth]{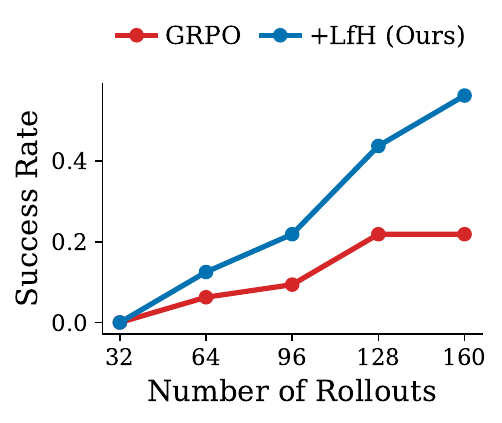}
    \caption{Real-world experimental results (evaluation success rate over training).}
    \label{fig:realworld}
    \vspace{-6pt}
\end{wrapfigure}

\textbf{Results.}
Figure~\ref{fig:realworld} shows that LfH improves real-world sample efficiency. Both methods start from zero, but LfH improves faster. At 128 rollouts, LfH achieves roughly twice the success rate of GRPO; at 160 rollouts, LfH reaches $56\%$ success, while GRPO plateaus at $22\%$.

\textbf{Analysis.}
The initial policy often grounds instructions to the wrong object, for example grasping the tape regardless of the commanded instruction. GRPO treats these rollouts only as failures. LfH instead relabels them with instructions describing the executed behavior, such as ``pick up the tape'' or ``move the tape to the bowl,'' turning failed target-task attempts into useful supervision for object grounding and manipulation.

\raggedbottom
\section{Discussion and Conclusion}
\label{sec:conclusion}

\textbf{More bits per robot sample.}
LfH is motivated by a simple question: how many useful bits do we extract from each costly robot rollout? Classic RL uses a rollout mainly to evaluate whether the intended outcome happened, often reducing a rich trajectory to one reward. This is inefficient when physical interaction is expensive. Following the broader view that learning systems should extract more information from each sample \cite{lecun2022path}, LfH treats a trajectory as evidence for multiple language goals consistent with what happened. In this sense, hindsight relabeling increases the bits per sample: one rollout can supervise multiple possible tasks, rather than only the task originally commanded.

\textbf{Practical implication.}
This is especially natural for pretrained VLAs. Because these models already generalize across related language instructions, training on what the robot actually did can improve what it was asked to do. LfH therefore changes the role of robot data: failed attempts are no longer merely negative examples, but reusable experience that can be mined for nearby tasks. This offers a practical path for VLA post-training: instead of only collecting more rollouts or taking more optimization steps on the same labels, we can reinterpret each rollout through language and transfer learning across the tasks it supports.

\textbf{Limitations.}
LfH is limited by the behaviors present in the collected data. If the policy produces repetitive or uninformative trajectories, there are few meaningful hindsight goals to recover. It also depends on relabeling quality, since incorrect or overly generic instructions introduce noisy supervision. Better filtering, uncertainty estimation, and lightweight human verification are important directions for reliable deployment.

\section*{Author Contributions}

\begin{itemize}
    \item \textbf{Iris Xu:} Led the project, refined and implemented the method, conducted simulation and physical robot experiments, and drafted the manuscript.
    \item \textbf{Sunshine Jiang:} Conducted simulation experiments on LIBERO-PRO position perturbations and physical robot experiments, and helped collect demonstrations for the latter.
    \item \textbf{John Marangola:} Co-implemented the training and inference pipeline for the physical robot experiments and helped collect the corresponding demonstrations.
    \item \textbf{Pulkit Agrawal:} Advised on the paper's framing and positioning.
    \item \textbf{Zhang-Wei Hong:} Conceived the idea, coordinated the research team, developed an early prototype of the method, and co-wrote the manuscript.
\end{itemize}

\acknowledgments{We thank Nitish Dashora, Richard Li, Thomas Liu, Aldo Pareja, Zexue He, Yuheng Zhi, and Alex Pentland for their feedback and comments to this work. John Marangola was financially supported by the Ministry of Trade, Industry, and Energy (MOTIE), Korea, under the “Global Industrial Technology Cooperation Center program” supervised by the Korea Institute for Advancement of Technology (KIAT). (Grant No. P0028435)}

\setlength{\bibsep}{0pt}
\bibliography{main}
\clearpage
\appendix
\section{Extended Experimental Results}
\label{sec:additional_results}

\subsection{Additional ablation studies}
\label{app:ablation}

\begin{figure}[H]
    \centering

    \begin{subfigure}[b]{0.49\linewidth}
        \centering
        \includegraphics[width=\linewidth]{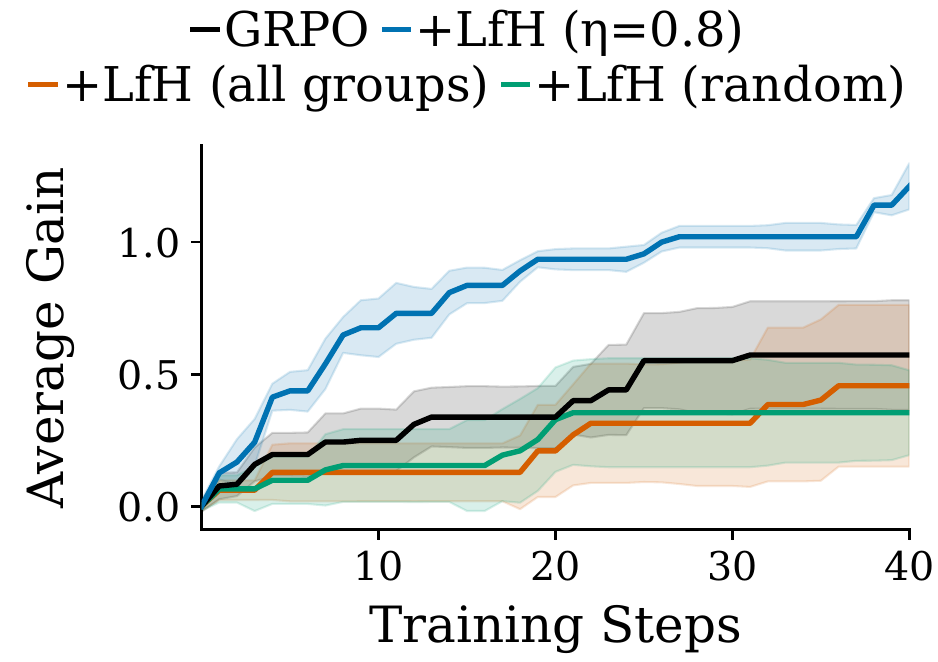}
        \caption{}
        \label{fig:group-selection-selection}
    \end{subfigure}
    \hfill
    \begin{subfigure}[b]{0.49\linewidth}
        \centering
        \includegraphics[width=\linewidth]{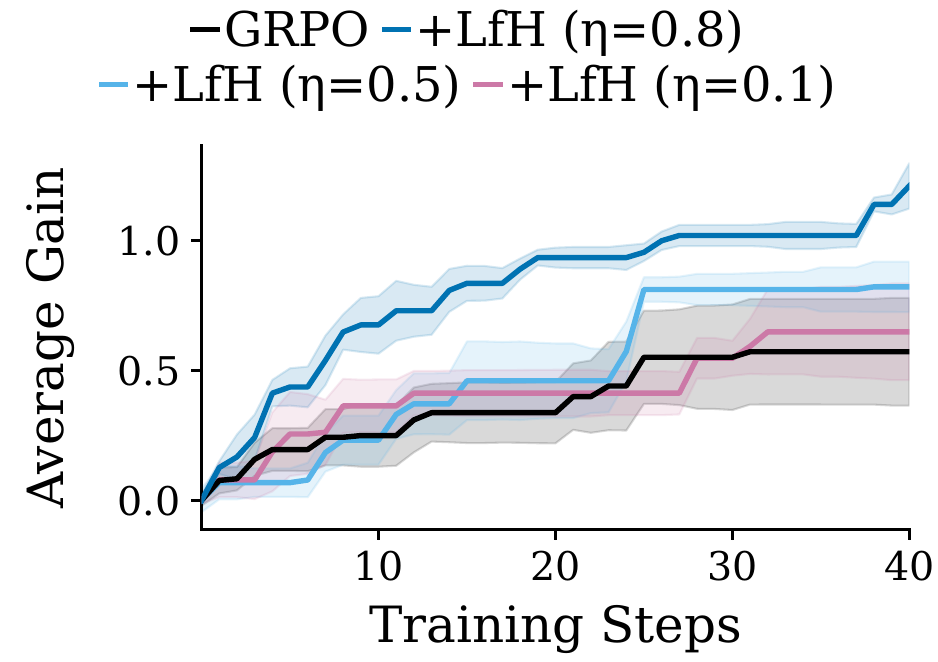}
        \caption{}
        \label{fig:group-selection-threshold}
    \end{subfigure}

    \par\medskip

    \begin{subfigure}[b]{0.49\linewidth}
        \centering
        \includegraphics[width=\linewidth]{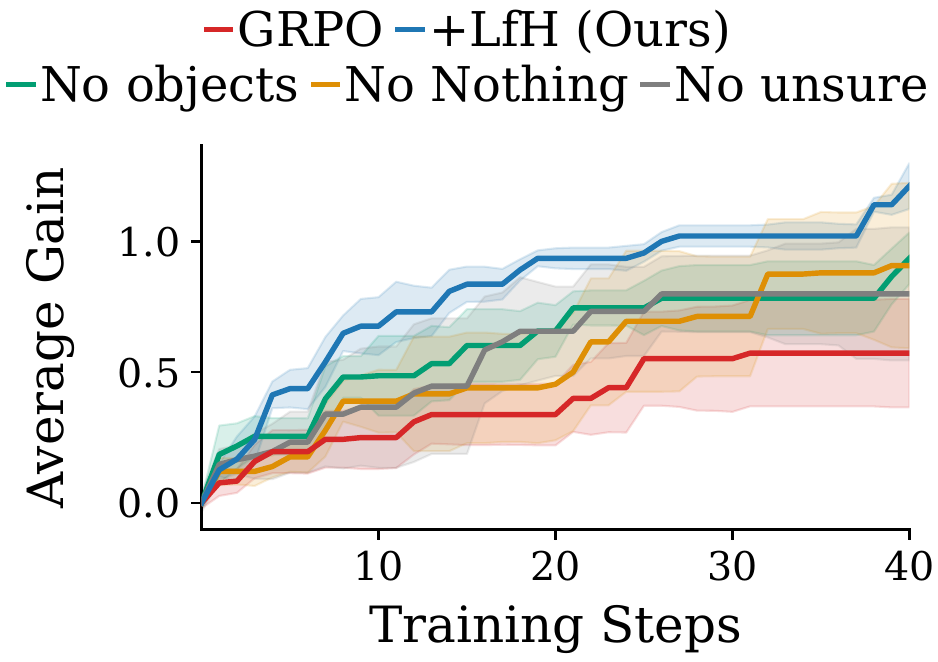}
        \caption{}
        \label{fig:appendix_prompt_ablation}
    \end{subfigure}
    \hfill
    \begin{subfigure}[b]{0.49\linewidth}
        \centering
        \includegraphics[width=\linewidth]{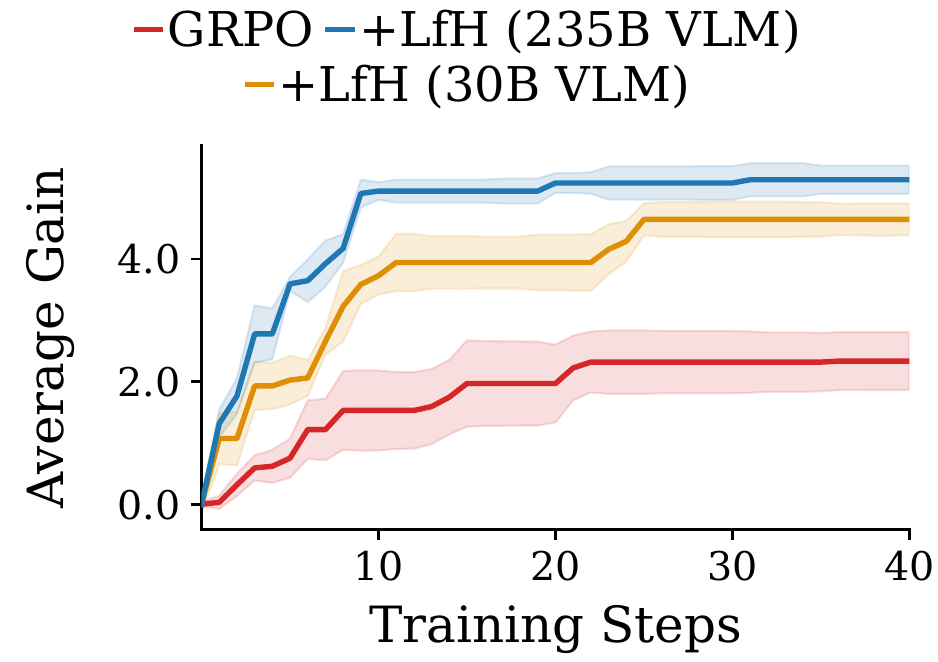}
        \caption{}
        \label{fig:weak_vlm_ablation}
    \end{subfigure}

    \caption{
        Further ablations.  \textbf{(a)} Selecting groups with mean reward below 0.8
        outperforms random selection and relabeling all groups,
        showing targeted relabeling is important.
        \textbf{(b)} Among the tested thresholds, 0.8 performs best.
        \textbf{(c)} Removing the scene object list,
        the \texttt{Nothing} option for filtering uninteresting
        trajectories, or the \texttt{unsure} option for reward
        evaluation each reduces the gains of LfH.
        \textbf{(d)} LfH with a weaker VLM (30B vs 235B) on LIBERO-PRO Object preserves most of LfH's gains.
    }
    \label{fig:appendix-ablations}
\end{figure}

Figure~\ref{fig:appendix-ablations} collects the group-selection, prompt, and VLM-size ablations discussed below.

\textbf{Relabeled group selection.} We relabel only groups whose mean original-task reward falls below the threshold $\eta=0.8$, unless otherwise specified. This targets groups with enough failures to yield useful hindsight supervision while avoiding the compute and noisy or redundant updates caused by relabeling successful groups.

Figure~\ref{fig:group-selection-selection} compares our 0.8 threshold with two variants: random selection of the same 70\% of groups and relabeling all groups. Thresholding performs substantially better than both. Random selection wastes VLM queries on uninformative groups, while relabeling all groups can dilute useful original-task supervision with redundant or noisy hindsight updates.

We also ablate the threshold used for filtering groups. As shown in Figure~\ref{fig:group-selection-threshold}, a very low threshold, such as 0.1, relabels only all-failed groups and therefore misses many trajectories that fail the commanded task but still contain meaningful, reusable object interactions. A more moderate threshold, such as 0.5, improves over this stricter setting but still underperforms the default threshold of 0.8. LfH thus benefits most from relabeling a broad set of non-saturated groups.

\textbf{VLM prompt.} LfH uses two structured templates. For instruction relabeling, the VLM receives a rollout and a list of scene objects, then returns a concise instruction describing the achieved behavior or \texttt{Nothing} if no meaningful interaction occurred. For reward relabeling, it receives the rollout and proposed hindsight instruction, then classifies completion as success, failure, or \texttt{unsure}, mapped to rewards 1, 0, and 0.5, respectively. On LIBERO-PRO Goal with task perturbation using $\pi_{0.5}$, we ablate the object list and the two abstention options. Figure~\ref{fig:appendix_prompt_ablation} shows that removing any component reduces sample efficiency. Without the object list, labels become less grounded and consistent; without \texttt{Nothing}, idle or contact-free rollouts become noisy positive examples; without \texttt{unsure}, ambiguous videos receive potentially incorrect binary rewards. Thus, effective relabeling requires prompts that constrain labels and permit filtering or abstention.

\textbf{Weaker VLM.} We also evaluate LfH with Qwen3-VL-30B-A3B-Thinking: it retains most of the 235B VLM's gain and clearly outperforms GRPO (Figure~\ref{fig:weak_vlm_ablation}).

\begin{figure}[!t]
    \centering

    \begin{subfigure}[t]{0.49\linewidth}
        \centering
        \includegraphics[width=\linewidth]{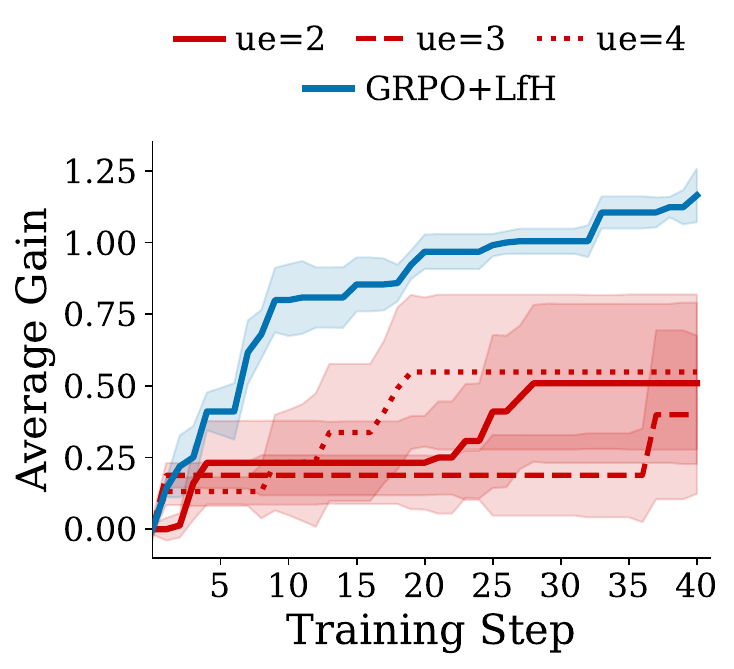}
        \caption{LfH vs GRPO}
        \label{fig:ue_sweep_grpo}
    \end{subfigure}
    \begin{subfigure}[t]{0.49\linewidth}
        \centering
        \includegraphics[width=\linewidth]{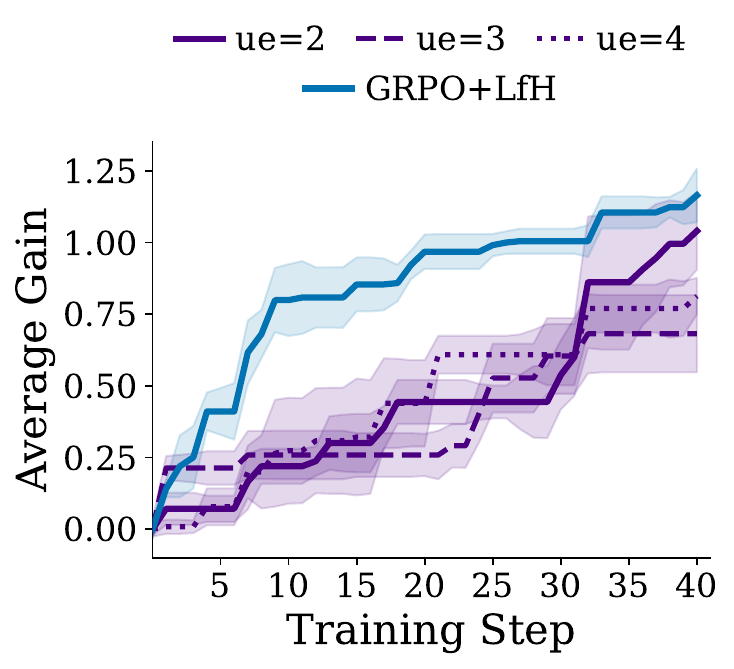}
        \caption{LfH vs PPO}
        \label{fig:ue_sweep_ppo}
    \end{subfigure}

    \caption{\textbf{(a)} More GRPO update epochs (ue) do not match LfH, suggesting LfH works by extracting more information per trajectory, rather than merely reusing data more aggressively. \textbf{(b)} More PPO epochs also fail to consistently match LfH.}
    \label{fig:ue_sweep}
\end{figure}

Figure~\ref{fig:ue_sweep} compares update epochs for GRPO (panel~\subref{fig:ue_sweep_grpo}) and PPO (panel~\subref{fig:ue_sweep_ppo}).

\begin{figure}[!t]
    \centering
    \includegraphics[width=0.8\linewidth]{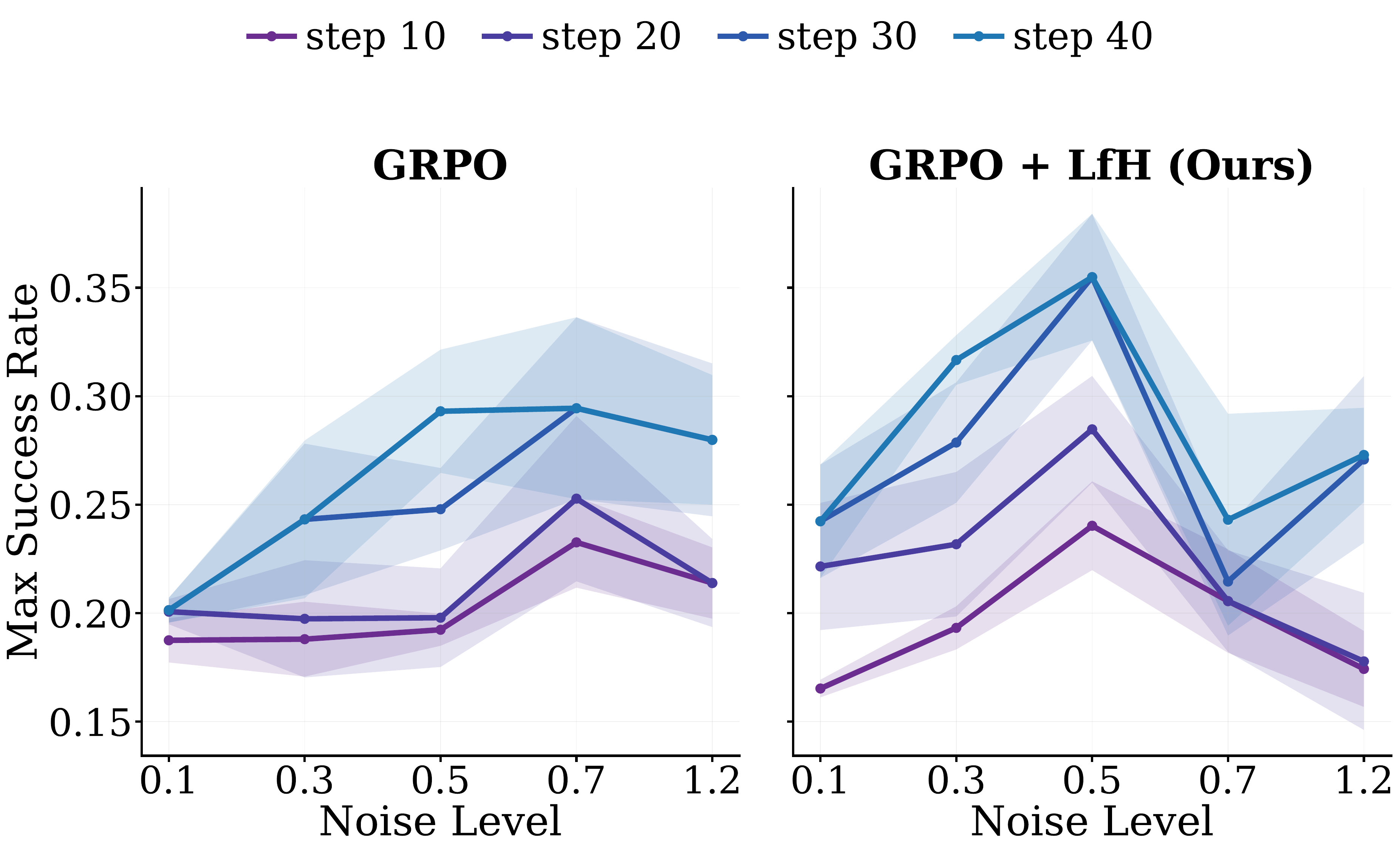}
    \caption{Ablation of the noise level used for rollout sampling. LfH is more sensitive to noise and benefits more from a more diverse rollout distribution.}
    \label{fig:noise_ablation}
\end{figure}

\subsection{Additional analysis of LfH}
\label{app:additional_analysis}

\begin{figure}[H]
    \hfill
    \centering
    \begin{subfigure}[t]{0.48\linewidth}
        \centering
        \includegraphics[width=\linewidth]{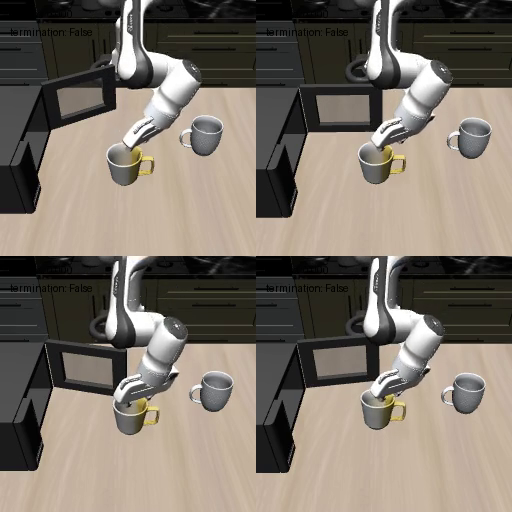}
        \caption{Repeated GRPO failures}
        \label{fig:diversity_rollout_grpo}
    \end{subfigure}
    \hfill
    \begin{subfigure}[t]{0.48\linewidth}
        \centering
        \includegraphics[width=\linewidth]{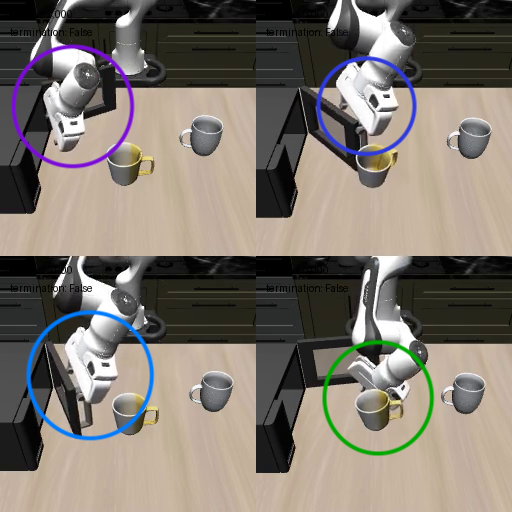}
        \caption{Diverse LfH rollouts}
        \label{fig:diversity_rollout_lfh}
    \end{subfigure}

    \caption{Random rollouts at training step 40 on the microwave task for \textbf{(a)} GRPO and \textbf{(b)} LfH. Standard GRPO collapses toward repeated, low-diversity failure modes, while LfH maintains more diverse and semantically meaningful behavior.}
    
    \label{fig:diversity_rollouts}
\end{figure}

Figure~\ref{fig:diversity_rollouts} contrasts the microwave rollouts under GRPO (panel~\subref{fig:diversity_rollout_grpo}) and LfH (panel~\subref{fig:diversity_rollout_lfh}).

\textbf{Hindsight correction diagnostics.} Equation~\eqref{eq:lfh_ratio} trades the unbiasedness of full HPG for much lower variance over long VLA horizons. Empirically, the median local ratio is $0.94$, whereas the median cumulative HPG ratio is $1.9\times10^{-12}$; $16.6\%$ of cumulative ratios underflow to zero, and the largest $10\%$ carry $68.7\%$ of the total weight. Both literal HPG and a self-normalized variant underperform standard GRPO.

\textbf{Can LfH improve sample efficiency on other OOD perturbations?}
We next evaluate a less obvious setting for language-based hindsight: position perturbations. Here, the commanded language prompt is unchanged, but the initial object positions are shifted. Since LfH relabels trajectories in language space, it is not immediately clear that it should help when only the environment configuration changes. Surprisingly, the preliminary results in Figure~\ref{fig:appendix_training_curves_pi05_position} suggest that LfH can still improve over GRPO on position-perturbation tasks, both in sample efficiency and maximum success rate. This points to a possible broader benefit of LfH: relabeled failures may help the policy reuse experience from shifted layouts and learn object interactions that can transfer back to the commanded task.

\subsection{Additional mechanism and robustness analyses}
\label{app:mechanism_analysis}

\textbf{Instruction--behavior pairing.}
In a follow-up analysis of the converged microwave policy, mug-directed behavior under ``close the microwave'' dropped to $0\%$ as task success rose from $0\%$ to $94\%$. We also permuted hindsight instructions across rollout groups while keeping the trajectories, rewards, and advantages fixed. Microwave success fell from $94\%$ to $0\%$, indicating that the instruction--behavior pairing matters beyond introducing reward variation.

\begin{figure}[H]
    \centering
    \begin{subfigure}[t]{0.48\linewidth}
        \centering
        \includegraphics[width=\linewidth]
        {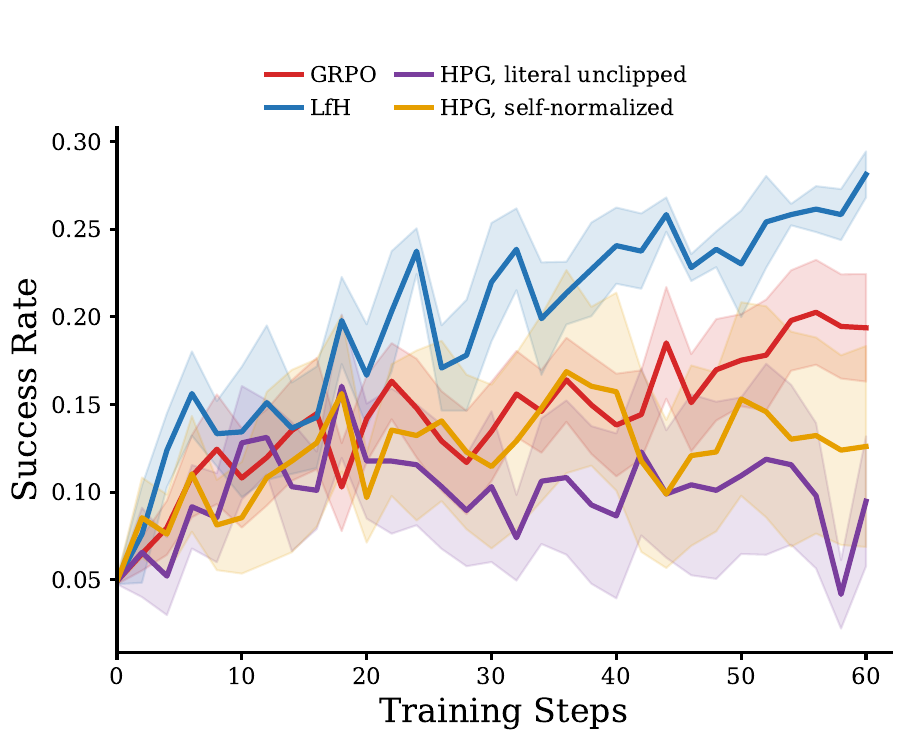}
        \caption{Per-decision HPG comparison.}
        \label{fig:hpg_training}
    \end{subfigure}
    \hfill
    \begin{subfigure}[t]{0.48\linewidth}
        \centering
        \includegraphics[width=\linewidth]
        {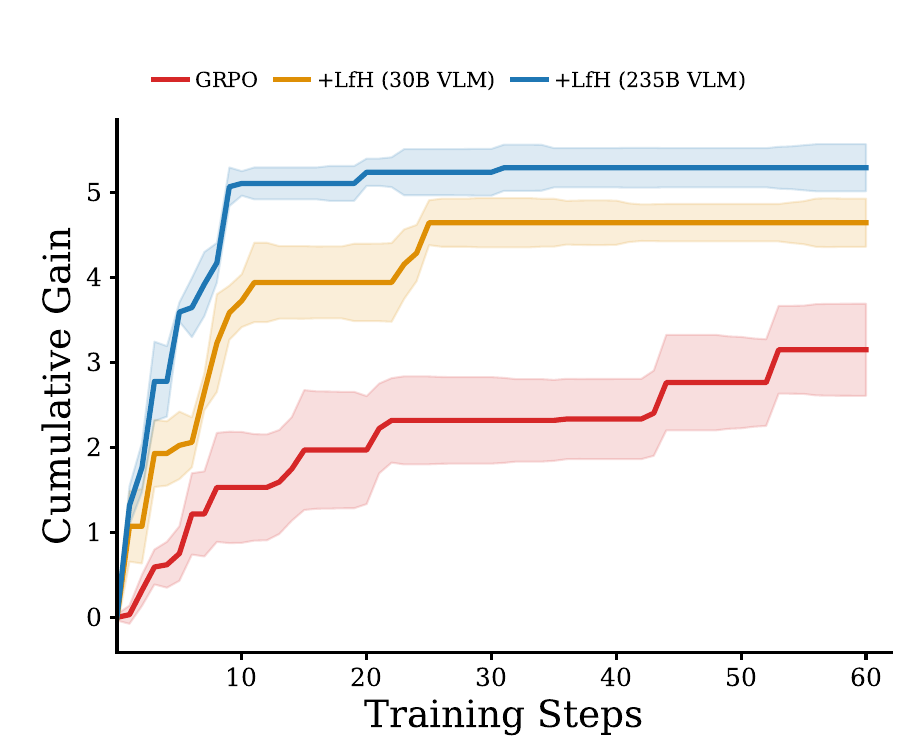}
        \caption{LfH with different VLM sizes.}
        \label{fig:weak_vlm_training}
    \end{subfigure}
    \caption{Additional analyses on LIBERO-PRO Object.}
    \label{fig:hpg_vlm_ablations}
    \vspace{-8pt}
\end{figure}

\textbf{Hindsight correction and VLM size.}
The single-step cross-instruction ratio avoids the weight degeneration of cumulative per-decision HPG~\cite{rauber2017hindsight}: $16.6\%$ of cumulative ratios underflow to zero, and the largest $10\%$ carry $68.7\%$ of total weight. Unclipped HPG and a self-normalized variant both underperform GRPO, while LfH remains stable. The smaller Qwen3-VL-30B-A3B-Thinking relabeler retains most of the 235B model's gain and outperforms GRPO. Figure~\ref{fig:hpg_vlm_ablations} shows these comparisons: panel~\subref{fig:hpg_training} compares HPG estimators, and panel~\subref{fig:weak_vlm_training} compares VLM sizes.

\textbf{Language-relabeling baseline.}
We adapted Hindsight Instruction Relabeling (HIR)~\cite{zhang2023wisdom} to the online LIBERO-PRO Goal setting, imitating each relabeled trajectory with a contrastive loss. By step 20, HIR reached $1.5\!\pm\!0.6\%$, compared with $22.8\!\pm\!2.6\%$ for GRPO and $29.1\!\pm\!1.3\%$ for LfH.

\textbf{Cross-suite transfer.}
Policies trained on LIBERO-PRO Object were evaluated on LIBERO-PRO Spatial instructions with unseen tasks and scenes. LfH performed better on five of eight instructions, tied on two, and performed worse on one.

\subsection{Full suite-wise validation curves}
\begin{figure}[H]
    \centering
    \includegraphics[width=\linewidth]{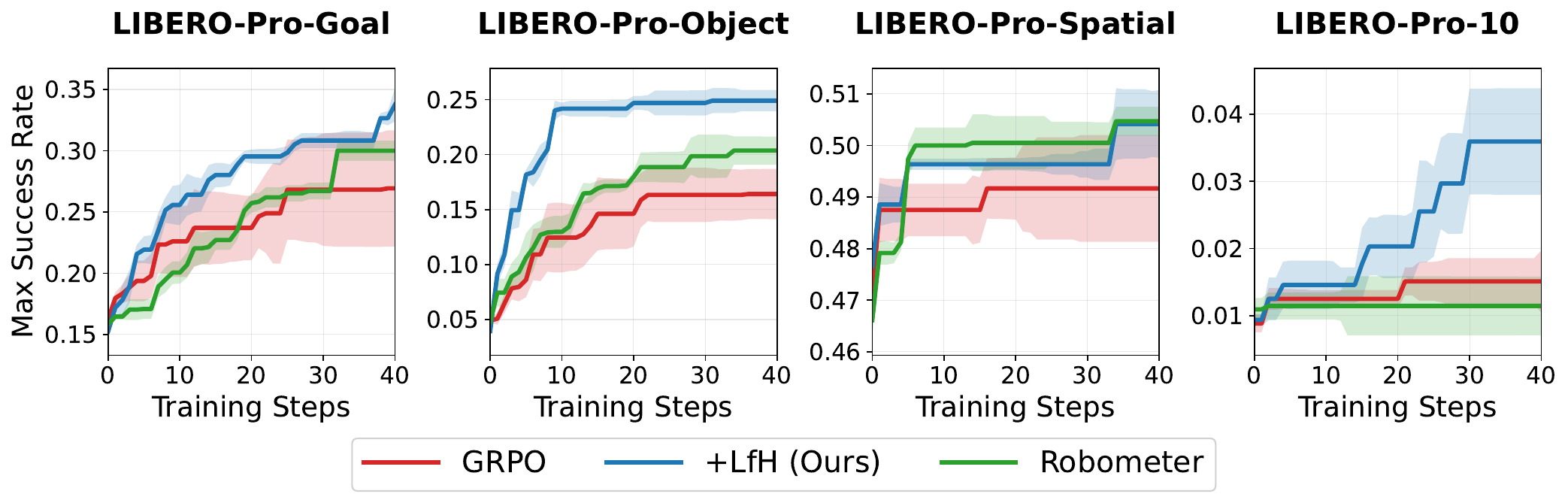}
    \caption{Best validation performance up to each training step for GRPO, GRPO+RoboMETER, and GRPO+LfH. Lines and shading denote the mean and standard error over four seeds.}
    \label{fig:full_training_curves}
\end{figure}

We report the cumulative maximum validation performance of $\pi_{0.5}$ on each LIBERO-PRO suite in Figure~\ref{fig:full_training_curves}.

\subsection{Full training curves}
\label{app:full_training_curves}

\begin{figure}[H]
    \centering
    \includegraphics[width=\linewidth]{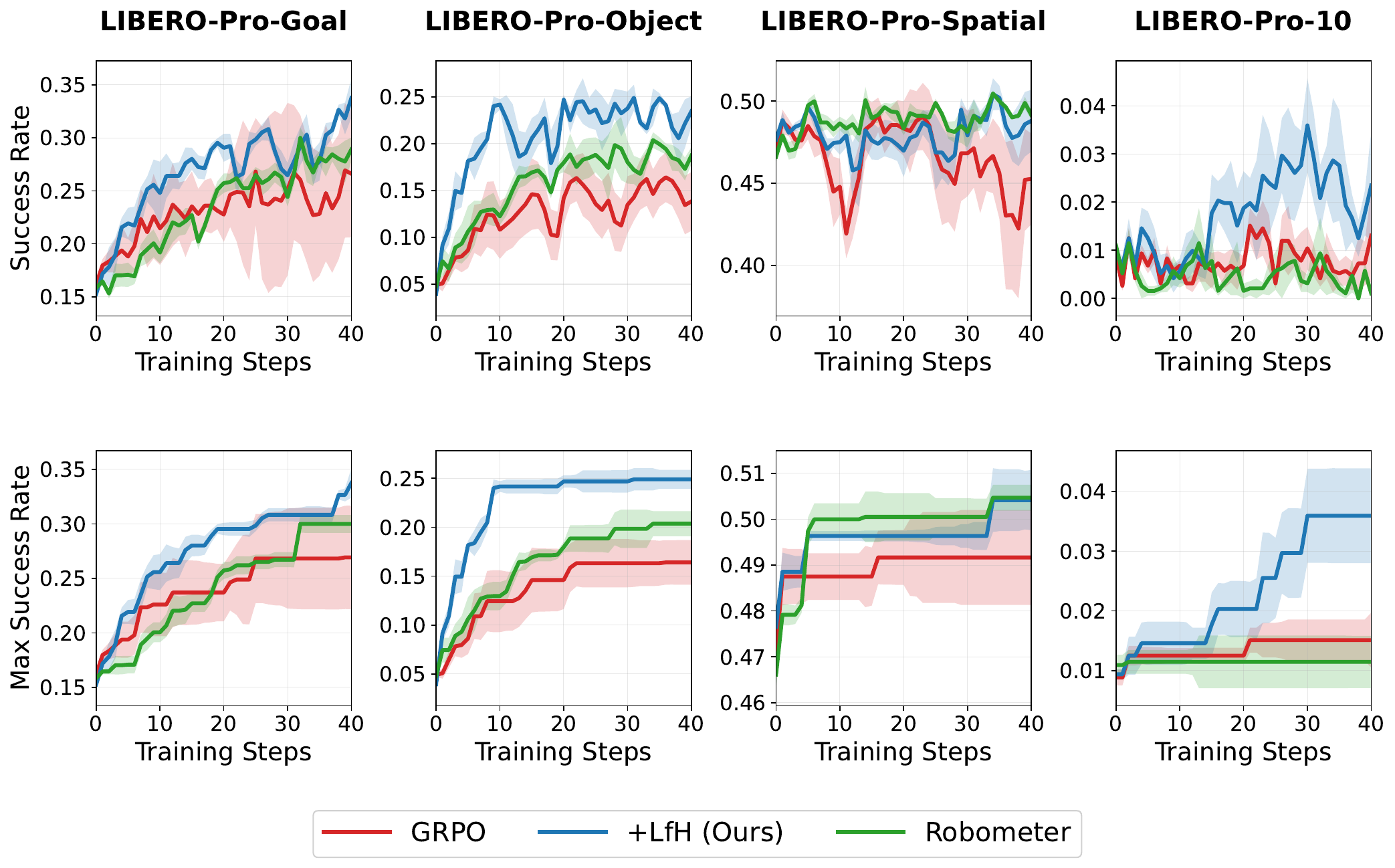}
    \caption{Full training curves for $\pi_{0.5}$.}
        \label{fig:appendix_training_curves_pi05}
\end{figure}

\begin{figure}[H]
    \centering
    \includegraphics[width=\linewidth]
    {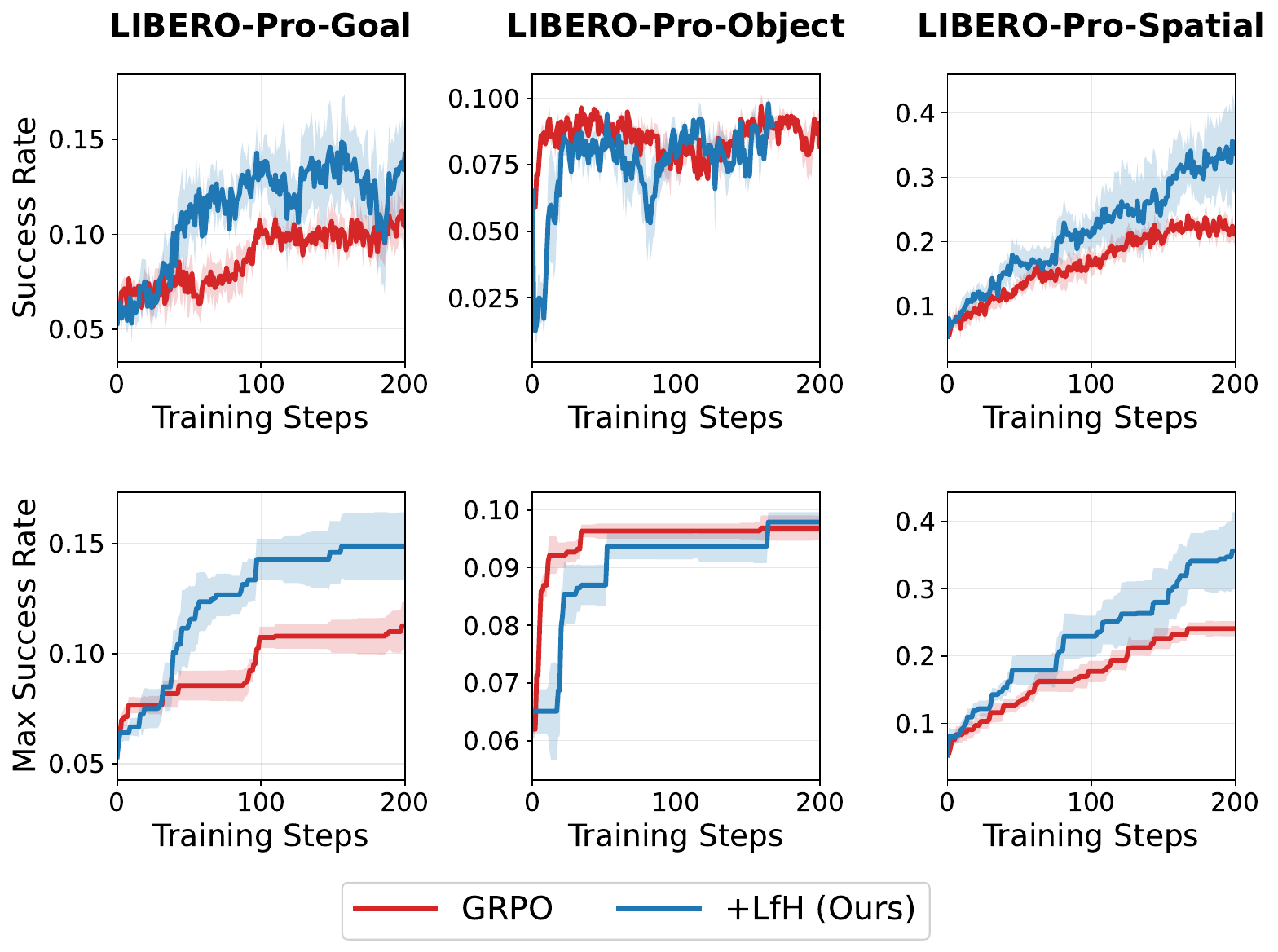}
    \caption{Full training curves for GR00T.}
        \label{fig:appendix_training_curves_gr00t}
\end{figure}

\begin{figure}[H]
    \centering
    \includegraphics[width=0.9\linewidth]{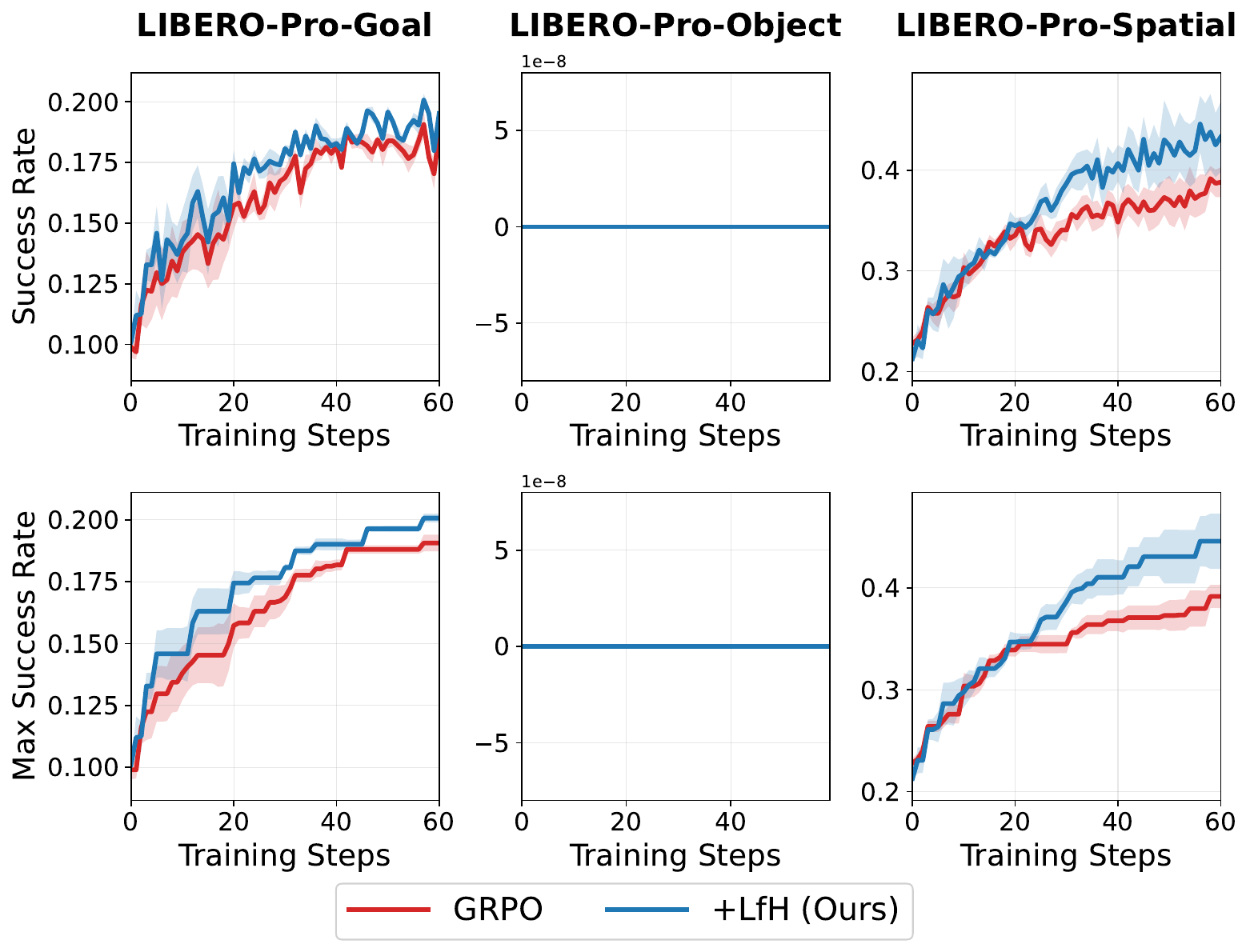}
    \caption{Full training curves for OpenVLA-OFT.}
    \label{fig:appendix_training_curves_openvla}
\end{figure}

\begin{figure}[H]
    \centering
    \includegraphics[width=0.9\linewidth]{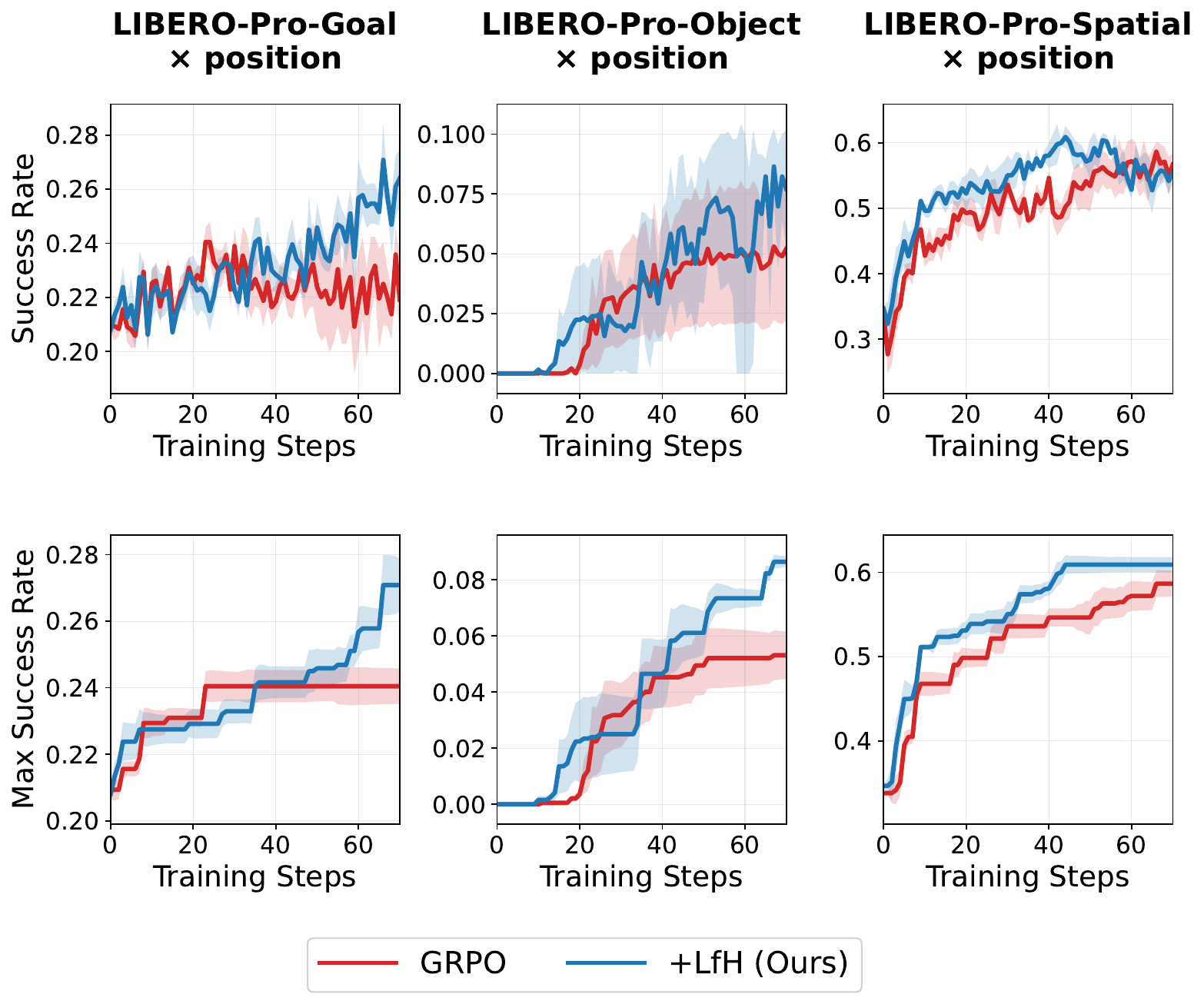}
    \caption{Full training curves for $\pi_{0.5}$ on position perturbation.}
        \label{fig:appendix_training_curves_pi05_position}
\end{figure}

The main metric is the gain defined in Section~\ref{subsec:exp:sample_efficiency}, averaged across suites. Here we also report raw success rates and cumulative maximum success rates for each model and suite. Figures~\ref{fig:appendix_training_curves_pi05}, \ref{fig:appendix_training_curves_gr00t}, and~\ref{fig:appendix_training_curves_openvla} show the $\pi_{0.5}$, GR00T, and OpenVLA-OFT results, respectively. Figure~\ref{fig:appendix_training_curves_pi05_position} shows the position-perturbation results.

\clearpage
\section{Implementation details}
\label{app:impl}

\subsection{LfH experimental details}
\paragraph{Hyperparameters.}
For the 64-rollout simulation experiments, we use $64$ environments with one rollout collection per step. We use group size $K=8$, microbatch $32$, global batch $512$, $\mathrm{KL\_beta}=0$ (no KL penalty), two gradient update epochs, and VLM relabeling temperature $0.8$. 

\paragraph{Prompts.}
\label{app:lfh_vlm_prompts}
LfH uses the VLM to both relabel the prompt and also assign rewards. The relabeling prompt is:
\begin{tcolorbox}[
    title=Relabeling Prompt,
    colback=gray!5,
    colframe=gray,
    boxrule=0.4pt,
    arc=1.5mm,
    toptitle=1mm,
    breakable
]

\scriptsize
\ttfamily
Watch this video of a robot arm performing a manipulation task.

The robot was originally instructed to:

"\{original\_instruction\}"

However, the robot may have FAILED or done something DIFFERENT from the original instruction.

Your job is to describe what the robot ACTUALLY DID based solely on visual evidence.

Objects present in the scene:

\{scene\_objects\_text\}

First, determine whether the robot did anything INTERESTING.

Interesting behavior means the robot PHYSICALLY CONTACTED an object --- the gripper fingers visibly touched or closed around it.

Uninteresting behavior means the robot merely hovered near an object, approached without making contact, or stayed idle.

CONTACT CHECK --- before claiming the robot touched anything, ask yourself:

- Did the gripper fingers visibly close around or press against the object?
- Did the object move, tilt, or shift in response to the gripper?
- Did the object's shadow change?

If NONE of these are true, the robot did NOT contact the object --- even if the gripper moved close to it.

If the behavior is UNINTERESTING (no confirmed physical contact), output exactly:

<thought>brief reasoning about why nothing interesting happened</thought>

<final>Nothing</final>

If the behavior IS interesting (confirmed contact), focus on:

1. Which object(s) did the gripper physically contact?
2. Where did the object(s) move to?
3. What is the final spatial arrangement?

Rules:

- Write ONE imperative instruction (e.g., "pick up the mug and place it in the microwave").
- Use SHORT, GENERIC names for objects: strip any numeric suffixes (\_1, \_2) and brand/material prefixes (e.g. "akita\_black\_bowl\_1" → "bowl", "flat\_stove\_1" → "stove").
- Do NOT copy the original instruction --- describe what you SEE.
- Under 20 words.

Output:

<thought>brief reasoning about what happened</thought>

<final>your instruction</final>
\end{tcolorbox}

The reward prompt is:

\begin{tcolorbox}[
    title=Reward Evaluation Prompt,
    colback=gray!5,
    colframe=gray,
    boxrule=0.4pt,
    arc=1.5mm,
    toptitle=1mm,
    breakable
]

\scriptsize
\ttfamily
Watch this video of a robot arm performing a manipulation task.

The robot was instructed to:

"\{instruction\}"

Determine whether the robot FULLY and SUCCESSFULLY completed the instruction.

Follow these steps:

1. FIRST FRAME: Describe the positions of all relevant objects at the start.
2. LAST FRAME: Describe the positions of all relevant objects at the end.
3. CHANGES: What specifically changed between the first and last frames? List each object that moved and where it moved to.
4. VERIFICATION: For each requirement in the instruction, state whether it is satisfied based on what you can clearly see in the last frame.
5. JUDGMENT: Are ALL requirements satisfied?

Rules:

- Only describe what you can clearly see in the frames. Do NOT infer or assume actions that are not visible --- narrating a plausible sequence of events is not evidence of completion.
- The task is complete ONLY if the last frame shows the EXACT goal state: correct objects in the correct locations as specified in the instruction.
- "In" a compartment/container means INSIDE the enclosed space --- not on top, not adjacent, not leaning against it.
- Moving an object to the WRONG location is a failure (e.g. wrong compartment, wrong side of the table), even if the object was moved.
- The robot arm may still be touching the object --- what matters is whether the object is at the target location.
- If you cannot clearly verify the target location in the last frame, answer UNSURE.

Output format must be exactly:

<thought>your step-by-step analysis following steps 1-5 above</thought>

<answer>yes</answer> or <answer>no</answer> or <answer>unsure</answer>

Use YES if the goal state is clearly achieved.

Use NO if it clearly is not achieved.

Use UNSURE if the evidence is ambiguous (e.g. you cannot tell which compartment an object is in, or whether it is on top vs inside).
\end{tcolorbox}

\paragraph{LfH group selection.}
Trajectories are grouped into sets of $K{=}8$. Groups whose mean reward falls within $[\theta_\text{low}, \theta_\text{high}]=[0.8, 1.0]$ are skipped; all other groups (mean reward $< 0.8$) are selected for hindsight relabeling. This focuses relabeling on groups with substantial failure rates where hindsight instructions provide the most value, while leaving near-saturated groups untouched.      

\paragraph{Hindsight loss clipping.}
LfH can occasionally produce very large hindsight-policy losses, especially early in training when hindsight rewards are noisy. To prevent a small number of hindsight groups from dominating optimization, we clip the hindsight GRPO loss before applying the hindsight weight:
\[
L_{\mathrm{LfH}}(\theta)
=
L_{\mathrm{GRPO}}(\theta)
+
\lambda \cdot \min\left(L_{\mathrm{H\text{-}GRPO}}(\theta), c\right),
\]
where \(c\) is a clipping threshold and \(\lambda\) controls the overall contribution of hindsight updates. This bounds the maximum contribution of the hindsight term to \(\lambda c\), while leaving the standard GRPO objective unchanged.

We keep this maximum hindsight contribution fixed across all experiments. To set the clipping threshold, we estimate the empirical distribution of hindsight losses during the early stage of training and choose \(c\) as approximately the 85th percentile. We then choose \(\lambda\) such that the weighted LfH batch loss is roughly equal to the original batch policy loss.

\paragraph{Same-instruction skip.}
Before reward evaluation, any trajectory whose relabeled instruction matches the original task (after case/punctuation normalization) reverts to its original instruction and environment reward, avoiding redundant VLM queries.

\paragraph{Policy likelihoods.}
For flow-matching policies ($\pi_{0.5}$ and GR00T), we follow the flow-to-SDE reformulation, modeling the $N$-step denoising trajectory as Gaussian transitions $\mathcal{N}(\mu_\theta(x_i,t_i,o,g),\sigma_i^2 I)$, with $\mu_\theta$ obtained from the predicted velocity, and sum their log densities. For OpenVLA-OFT, we sum the per-token categorical log probabilities of its discrete action tokens. In both cases the cross-instruction ratio rescores the same executed trajectory under $g'$, changing only the instruction conditioning.

\paragraph{VLM computation.}
VLM relabeling and reward evaluation took approximately $155$ seconds per training step on a separate node with eight H100 GPUs.

\clearpage
\subsection{Real-world experimental details}
\label{app:real_world_details}

\begin{figure}[H]
    \centering
    \includegraphics[width=0.56\linewidth]{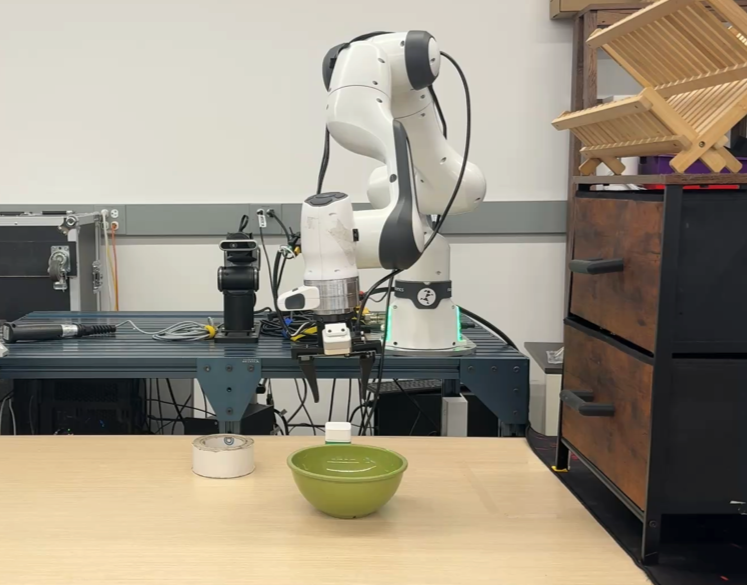}
    \caption{Real-world experimental setup.}
    \label{app:real_setup}
\end{figure}

\textbf{Setup.}
Our real-world setup consists of a Franka FR3 arm, a set of manipulable objects, two drawers, and a two-level rack. We first fine-tune $\pi_{0.5}$ with supervised learning (SFT) on 9 language-conditioned tasks that span different difficulty levels and horizon lengths, from short-horizon pick-and-place to multi-stage manipulation. For each task, we collect $20$ demonstration rollouts using SpaceMouse teleoperation. Figure~\ref{app:real_setup} shows the real-world setup. Table~\ref{app:tab:sft_tasks} lists the SFT task prompts.

\textbf{RL fine-tuning.}
We then compare GRPO and GRPO+LfH on the task \emph{put the green container in the bowl}, where the initial SFT policy achieves zero success. Rewards are provided by a human operator as binary labels. Each rollout has a time limit of $30$ seconds, corresponding to $512$ action steps; each step executes an action chunk of size $16$ predicted by the VLA model. For each of GRPO and GRPO+LfH, we use 32 rollouts per training step (160 total) and evaluate on separate rollouts after each training step. Each real-world method was evaluated with one training seed. Tables~\ref{tab:realworld-rl-hparams} and~\ref{tab:realworld-sft-hparams} list the GRPO and SFT hyperparameters, respectively.

\begin{table}[H]
\centering
\caption{Tasks used for collecting SFT demonstrations.}
\label{app:tab:sft_tasks}
\begin{tabular}{c l}
\toprule
\textbf{Task} & \textbf{Prompt} \\
\midrule
1  & put the tape on the top rack \\
2  & put the bowl on the top rack \\
3  & put the green container in front of the camera \\
4  & put the bowl in front of the camera \\
5  & open the top drawer and put the bowl inside \\
6  & open the bottom drawer and put the bowl inside \\
7  & open the top drawer and put the tape inside \\
8  & open the top drawer and put the green container inside \\
9  & put the green container on the bottom rack \\
\bottomrule
\end{tabular}
\end{table}

\begin{table}[H]
\centering
\caption{Real-world GRPO hyperparameters.}
\label{tab:realworld-rl-hparams}
\small
\begin{tabularx}{\columnwidth}{@{}lXr@{}}
\toprule
\textbf{Group} & \textbf{Hyperparameter} & \textbf{Value} \\
\midrule
\multirow{5}{*}{Rollout}
  & episodes per iteration         & 32 \\
  & max episode steps (chunks)     & 32 \\
  & poll interval (s)              & 60 \\
  & heartbeat interval (s)         & 60 \\
  & recompute logprobs             & False \\
\midrule
\multirow{3}{*}{Environment}
  & total num envs (train / eval)  & 32 / 32 \\
  & group size                     & 4 \\
  & max episode steps (robot)      & 512 \\
\midrule
\multirow{8}{*}{Algorithm}
  & loss aggregation               & token-mean \\
  & update epochs                  & 2 \\
  & rollout epoch                  & 1 \\
  & KL penalty type                & low\_var\_kl \\
  & KL coefficient $\beta$         & $1{\times}10^{-3}$ \\
  & entropy bonus                  & 0 \\
  & clip ratio (low / high)        & 0.2 / 0.2 \\
  & clip ratio $c$                 & 3.0 \\
\midrule
\multirow{4}{*}{Model}
  & action dim                     & 10 \\
  & num action chunks              & 16 \\
  & flow-matching steps            & 10 \\
  & noise level                    & 0.1 \\
\midrule
\multirow{6}{*}{Optimizer}
  & actor LR                       & $2{\times}10^{-6}$ \\
  & value LR                       & $1{\times}10^{-4}$ \\
  & Adam $(\beta_1, \beta_2)$      & $(0.9,\,0.95)$ \\
  & Adam $\varepsilon$             & $1{\times}10^{-8}$ \\
  & weight decay                   & 0.01 \\
  & grad clip                      & 1.0 \\
\midrule
\multirow{4}{*}{Batching / FSDP}
  & micro batch size               & 32 \\
  & global batch size              & 512 \\
  & sharding strategy              & \texttt{no\_shard} (DDP-style) \\
  & gradient checkpointing         & False \\
                        
\bottomrule
\end{tabularx}
\end{table}

\begin{table}[H]
\centering
\caption{Real-world SFT hyperparameters.}
\label{tab:realworld-sft-hparams}
\small
\begin{tabularx}{\columnwidth}{@{}lXr@{}}
\toprule
\textbf{Group} & \textbf{Hyperparameter} & \textbf{Value} \\
\midrule
\multirow{6}{*}{Model}
  & precision                      & bf16 \\
  & action dim                     & 10 \\
  & num action chunks              & 50 \\
  & num flow steps                 & 10 \\
  & num image inputs               & 2 (two cameras) \\
  & train expert only              & True \\
\midrule
\multirow{2}{*}{Batching}
  & micro batch size               & 1 \\
  & global batch size              & 256 \\
\midrule
\multirow{6}{*}{Optimizer}
  & actor LR                       & $1{\times}10^{-4}$ \\
  & value LR                       & $1.55{\times}10^{-4}$ \\
  & Adam $(\beta_1, \beta_2)$      & $(0.95,\,0.99)$ \\
  & Adam $\varepsilon$             & $1{\times}10^{-8}$ \\
  & weight decay                   & $1{\times}10^{-6}$ \\
  & grad clip                      & 1.0 \\
\midrule
\multirow{5}{*}{FSDP / mixed prec.}
  & strategy                       & FSDP \\
  & sharding strategy              & \texttt{no\_shard} \\
  & use orig params                & True \\
  & gradient checkpointing         & False \\
  & wrap policy                    & disabled \\
\midrule
\multirow{3}{*}{AMP}
  & enabled                        & True \\
  & precision                      & bf16 \\
  & grad scaler                    & False \\
\bottomrule
\end{tabularx}
\end{table}

\section{Additional examples of relabeled trajectories}
\label{app:examples}

\begin{figure}[H]
\centering
\scriptsize
\includegraphics[width=0.78\linewidth]{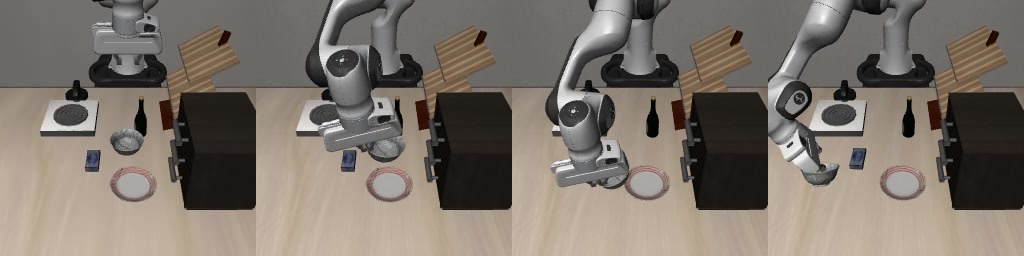}\par
\raggedright
\vspace{0.3em}
\quad \textbf{Original:} put the plate on the stove \\
\quad \textbf{Relabeled:} move the bowl to the left side of the table

\vspace{0.8em}
\hrule
\vspace{1em}

\centering
\includegraphics[width=0.78\linewidth]{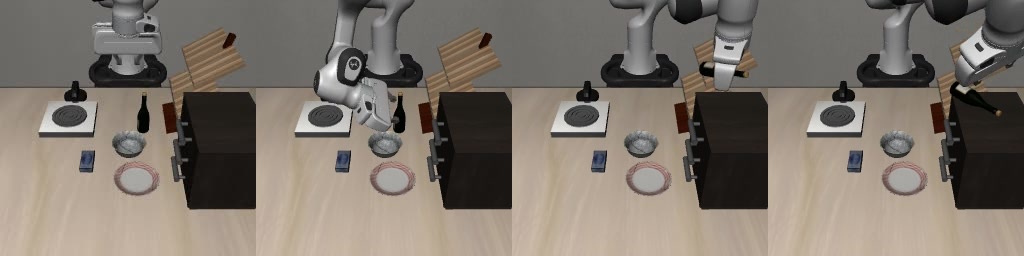}\par

\raggedright
\vspace{0.3em}
\quad \textbf{Original:} put the wine bottle on the plate \\
\quad \textbf{Relabeled:} place the wine bottle on the cabinet

\vspace{0.8em}
\hrule
\vspace{1em}

\centering
\includegraphics[width=0.78\linewidth]{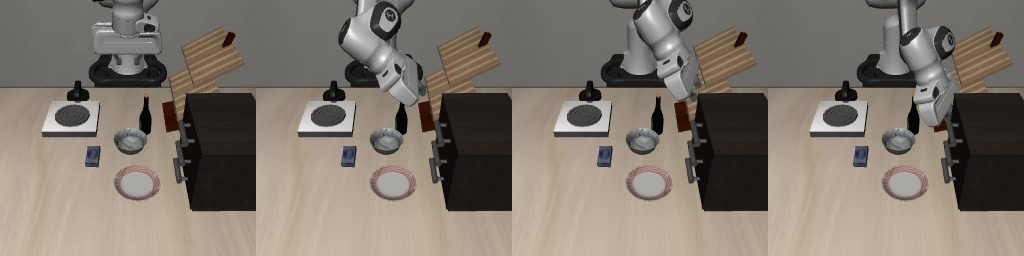}\par

\raggedright
\vspace{0.3em}
\quad \textbf{Original:} put the wine bottle on the plate \\
\quad \textbf{Relabeled:} nothing (uninteresting)
\vspace{0.8em}
\hrule
\vspace{1em}

\centering
\includegraphics[width=0.78\linewidth]{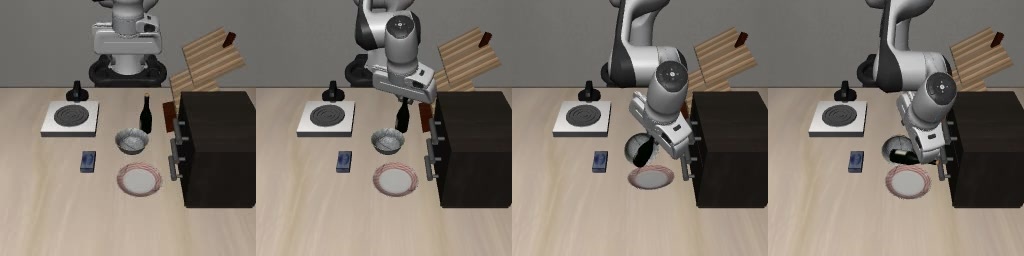}\par

\raggedright
\vspace{0.3em}
\quad \textbf{Original:} put the wine bottle in the bowl \\
\quad \textbf{Relabeled:} pour the wine bottle's contents into the bowl
\vspace{0.8em}
\hrule
\vspace{1em}
\caption{\textbf{Qualitative examples of LfH relabeling.} The VLM proposes an instruction describing the observed behavior; \texttt{Nothing} filters uninformative rollouts, though hallucinations remain possible.}
\label{fig:lfh_relabel_examples}
\end{figure}

Figure~\ref{fig:lfh_relabel_examples} shows relabeled trajectories, including a rollout filtered as uninteresting.

\end{document}